\pdfoutput=1
\documentclass[11pt]{article}

\usepackage[]{emnlp2021}

\usepackage{times}
\usepackage{latexsym}

\usepackage[T1]{fontenc}
\usepackage[utf8]{inputenc}

\usepackage{microtype}

\usepackage{tabularx}
\usepackage{booktabs}
\usepackage{tikz}
\usepackage{pgfplots}
\usepackage{amsmath}
\usepackage{amssymb}
\usepackage{graphicx}
\usepackage{multirow}
\usepackage{subcaption}
\DeclareMathOperator*{\argmax}{arg\,max}

\newcommand{\negphantom}[1]{\settowidth{\dimen0}{#1}\hspace*{-\dimen0}}

\newcommand{\peerEdit}[0]{PEER-Edit}
\newcommand{\peerUndo}[0]{PEER-Undo}
\newcommand{\peerExplain}[0]{PEER-Explain}
\newcommand{\peerDocument}[0]{PEER-Document}
\newcommand{\peerEditSynPlans}[1][]{PEER (SP{#1})}

\pgfplotsset{compat=1.13}

\definecolor{c0}{cmyk}{1,0.3968,0,0.2588} 
\definecolor{c1}{cmyk}{0,0.6175,0.8848,0.1490} 
\definecolor{c2}{cmyk}{0.1127,0.6690,0,0.4431} 
\definecolor{c3}{cmyk}{0.3081,0,0.7209,0.3255} 
\definecolor{c4}{cmyk}{0.6765,0.2017,0,0.0667} 
\definecolor{c5}{cmyk}{0,0.8765,0.7099,0.3647} 

\definecolor{c0alt}{RGB}{15,158,251} 

\definecolor{darkgrey}{RGB}{149,149,149}
\definecolor{decentgrey}{RGB}{242,242,242}

\usetikzlibrary{calc,fit,positioning,arrows,arrows.meta,backgrounds,decorations.pathreplacing}
\usepgfplotslibrary{fillbetween}

\title{PEER: A Collaborative Language Model}

\author{Timo Schick$^{\diamondsuit}$ \quad Jane Dwivedi-Yu$^{\diamondsuit}$ \quad Zhengbao Jiang$^{\diamondsuit,\heartsuit}$ \quad Fabio Petroni$^{\diamondsuit}$ \\[4pt] { \bf Patrick Lewis$^{\diamondsuit}$ \quad \bf Gautier Izacard$^{\diamondsuit,\clubsuit}$ \quad Qingfei You$^{\diamondsuit}$ \quad Christoforos Nalmpantis$^{\diamondsuit}$} \\[4pt] { \bf Edouard Grave$^{\diamondsuit}$\quad Sebastian Riedel$^{\diamondsuit,\spadesuit}$} \\[8pt]
$^{\diamondsuit}$ Meta AI Research,
$^{\heartsuit}$ Carnegie Mellon University,\\
$^{\clubsuit}$ Inria \& ENS, PSL University,
$^{\spadesuit}$ University College London \\[4pt]
{\tt \{schick,janeyu,zhengbao,fabiopetroni,plewis,gizacard}\\
{\tt qingfeiyou,christoforos,egrave,sriedel\}@fb.com}
}

\begin{document}
\maketitle

\begin{abstract}
    Textual content is often the output of a collaborative writing process: We start with an initial draft, ask for suggestions, and repeatedly make changes.
    Agnostic of this process, today’s language models are trained to generate only the final result. As a consequence, they lack several abilities crucial for collaborative writing: They are unable to update existing texts, difficult to control and incapable of verbally planning or explaining their actions.
    To address these shortcomings, we introduce PEER, a \emph{collaborative} language model that is trained to imitate the entire writing process itself: PEER can write drafts, add suggestions, propose edits and provide explanations for its actions. Crucially, we train multiple instances of PEER able to \emph{infill} various parts of the writing process, enabling the use of self-training techniques for increasing the quality, amount and diversity of training data. This unlocks PEER's full potential by making it applicable in domains for which no edit histories are available and improving its ability to follow instructions, to write useful comments, and to explain its actions. We show that PEER achieves strong performance across various domains and editing tasks.
\end{abstract}

\section{Introduction}

Large neural networks show impressive text generation capabilities when pretrained with a language modeling objective \citep[i.a.]{radford2018language,raffel2019exploring,brown2020language,rae2021gopher,zhang2022opt,chowdhery2022palm}. However, the way these models operate---producing outputs in a single pass from left to right---differs strongly from the iterative process by which humans typically write texts. This limits their utility for \emph{collaborative} writing in various respects; for example, they are not able to retroactively modify or refine their own outputs. Beyond that, they are hard to control \citep{pmlr-v162-korbak22a} and verifying their outputs is challenging as they often hallucinate content \citep{maynez-etal-2020-faithfulness,shuster-etal-2021-retrieval-augmentation,nakano2021webgpt} and lack the ability to explain their intentions. All of this makes it very difficult for humans to collaborate with such models for writing coherent, factual texts.

\begin{figure}
    \centering
    \includegraphics[width=\linewidth]{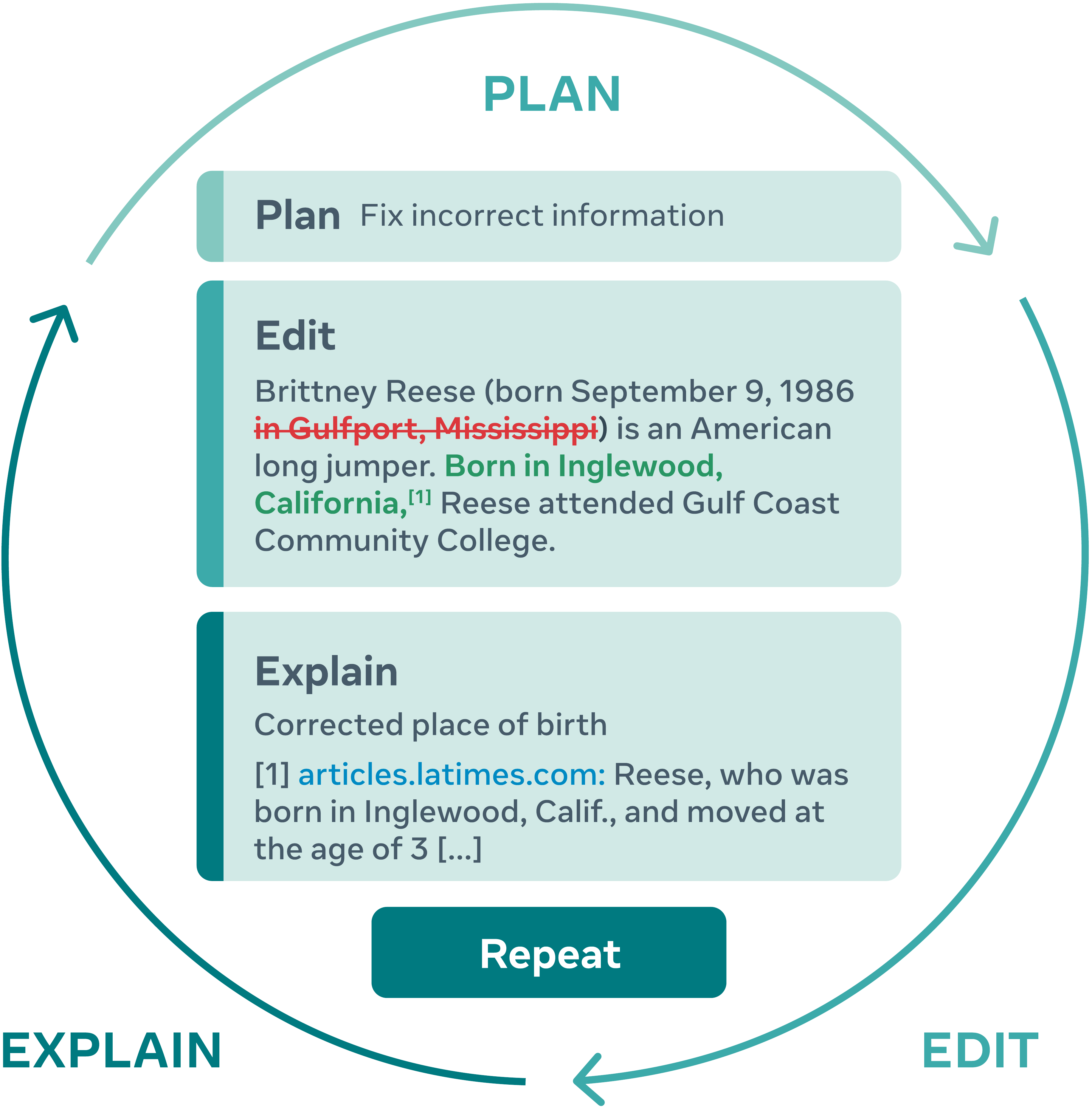}
    \caption{Illustration of the steps performed by PEER, our collaborative language model: First, either the user or the model specifies a \textbf{\emph{plan}} describing the action they want to be performed; this action is then realized by means of an \textbf{\emph{edit}}. The model can \textbf{\emph{explain}} the edit both in natural language and by pointing to relevant sources. We can \textbf{\emph{repeat}} this process until the generated text requires no further updates.}
    \label{fig:peer_idea}
\end{figure}

To address these shortcomings of existing LMs, we propose PEER~(\textbf{P}lan, \textbf{E}dit, \textbf{E}xplain, \textbf{R}epeat), a \emph{collaborative} language model trained on edit histories to cover the entire writing process.
As illustrated in Figure~\ref{fig:peer_idea}, PEER operates in several steps that aim to mirror the human writing process: For a given text, either a user or the model itself can \emph{plan} an action to be applied, for example by means of a natural language instruction. This plan is then realized by an \emph{edit}, which the model can \emph{explain} both in form of a textual comment and by pointing to references used; this is enabled by augmenting each input text with retrieved passages containing potentially relevant background information. We \emph{repeat} these steps until the text is in a satisfactory state that does not require any further updates. This iterative approach does not only enable the model to decompose the complex task of writing a consistent, factual text into multiple easier subtasks, it also allows humans to intervene at any time and steer the model in the right direction, either by providing it with their own plans and comments or by making edits themselves.

Similar to recent approaches for iterative editing~\citep{faltings-etal-2021-text,reid2022learning}, we use Wikipedia as our main source of edits and associated comments, which we use as proxies for plans and explanations. In contrast to this prior work, however, our goal is to obtain a collaborative model that is useful \emph{beyond} just Wikipedia: It should be capable of following human-written instructions for updating texts in any domain. To achieve this goal, we train PEER not only to perform the writing process illustrated in Figure~\ref{fig:peer_idea} in sequential order, but also to \emph{infill} various parts; for example, given an edited text and a set of relevant documents, we teach it to produce the original version of this text \emph{before} it was edited. This enables us to use self-training techniques~\citep[e.g.,][]{yarowsky-1995-unsupervised,sennrich-etal-2016-improving,He2020Revisiting,schick-schutze-2021-exploiting} for training PEER with synthetic plans, edits, explanations and documents. We show that this substantially improves PEER along several axes, including its ability to edit texts in any domain, to understand human-written instructions, and to explain its actions.

In summary, our contributions are as follows:

\begin{itemize}
\item We introduce PEER, a collaborative language model trained primarily on Wikipedia edit histories.
\item By training PEER to infill parts of the writing process and leveraging self-training techniques, we make it applicable in any domain and enhance several of its core capabilities essential for collaborative writing.
\item For different tasks related to editing texts, we show that PEER clearly outperforms various baselines and analyze factors leading to its strong performance.
\item To facilitate further research on collaborative LMs, we release a variety of PEER models as well as the data and code used to train them.
\end{itemize}

\section{Related Work}

\paragraph{Text Editing} Similar to our work, \citet{faltings-etal-2021-text} train an editing model to follow plans on Wikipedia data. However, they only consider single sentence edits, evaluate on Wikipedia data only and do not explore approaches for improving data quality and coverage. \citet{reid2022learning} also train models on Wikipedia's edit history, but do not consider plans, explanations or reference documents. Several editing models are trained to solve specific tasks, such as updating information~\citep{logan2021fruit}, fixing grammar errors~\citep{napoles2017jfleg,awasthi-etal-2019-parallel} or improving citations \citep{petroni-etal-2022-improving}. Various approaches teach models to iteratively improve texts in an unsupervised fashion~\citep[e.g.,][]{shen-etal-2020-blank,donahue-etal-2020-enabling,li2022diffusion} and explore more efficient ways of representing edits~\citep{mallinson-etal-2020-felix}. Closely related to our work, \citet{yang-etal-2017-identifying-semantic} classify edit intentions in Wikipedia's edit history.

\paragraph{Instruction Tuning and Planning} Explicitly teaching models to follow plans is closely related to recent work that finetunes models on large datasets of human-written instructions~\citep{wei2022finetuned,sanh2022multitask,bach-etal-2022-promptsource,ouyang2022training,wang2022benchmarking}. The idea of having a separate \emph{planning} stage has also been explored for other text generation tasks inlcuding summarization~\citep{10.1162/tacl_a_00438}, data-to-text generation~\citep{moryossef-etal-2019-step} and story writing \citep{Yao_Peng_Weischedel_Knight_Zhao_Yan_2019}. Our approach of writing coherent pieces of text by iteratively performing small updates has some similarity with recent approaches like \emph{chain-of-thought} prompting~\citep{wei2022chain,dohan2022language} and document sketching~\citep{wu-etal-2021-automatic}, that also break down a complex task into multiple smaller steps.

\paragraph{Collaborative Writing}  \citet{du-etal-2022-read,du2022understanding} investigate human-machine interactions for iteratively improving documents; however, they focus mostly on syntactic edits that improve the fluency, coherence or style of a document. \citet{lee2022coauthor} investigate using GPT3 \citep{brown2020language} as a writing assistant for creative and argumentative writing. In their setup, however, the model provides suggestions for continuations without being controllable by means of natural language instructions.

\paragraph{Self-Training} Our approach of using models to infill missing data and then train other models on this synthetic data closely resembles other self-training and bootstrapping approaches used e.g. in word sense disambiguation \citep{yarowsky-1995-unsupervised}, machine translation~\citep{sennrich-etal-2016-improving,hoang-etal-2018-iterative}, sequence generation~\citep{He2020Revisiting}, and few-shot learning~\citep{schick-schutze-2021-exploiting,schick-schutze-2021-shot}. Similar to how we use models to turn plain texts into sequences of edits, \citet{dai2022dialog} turn documents into dialogue sequences.

\section{Plan, Edit, Explain, Repeat}

\begin{figure*}[t!]
\begin{subfigure}{0.47\textwidth}
  \centering
  % include second image
  \includegraphics[width=\linewidth]{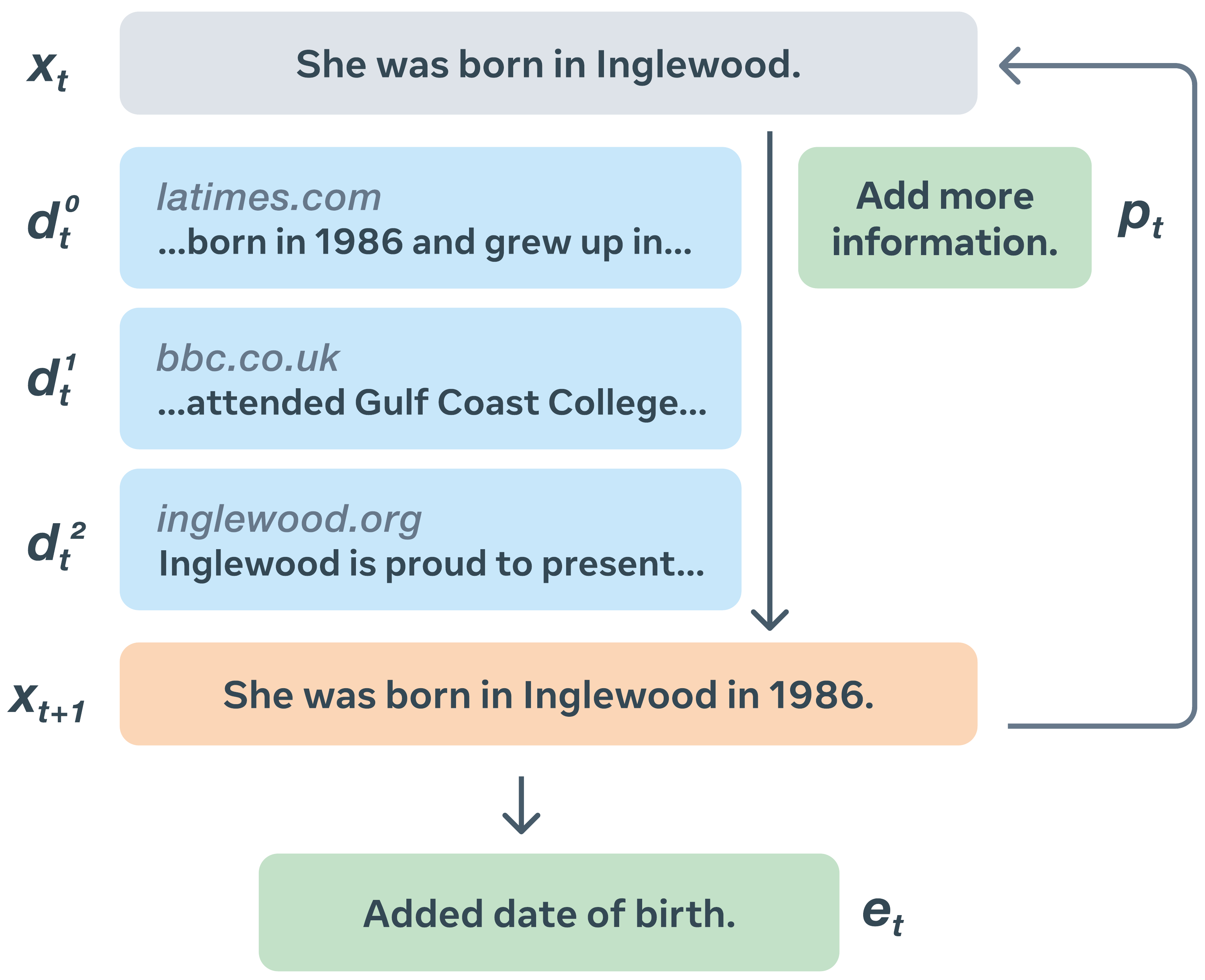}  
  \caption{Starting from a text $\mathbf{x}_t$, we use both a plan $\mathbf{p}_t$ and a collection of documents $D_t = \{ d_t^0, \ldots, d_t^k \}$ to obtain an updated version $\mathbf{x}_{t+1}$ and an explanation $\mathbf{e}_t$ of the performed edit; this process is repeated multiple times.}
  \label{fig:peer_process_a}
\end{subfigure}\hspace{0.02\textwidth}% Space between image a and b
\begin{subfigure}{0.47\textwidth}
  \centering
  \includegraphics[width=\linewidth]{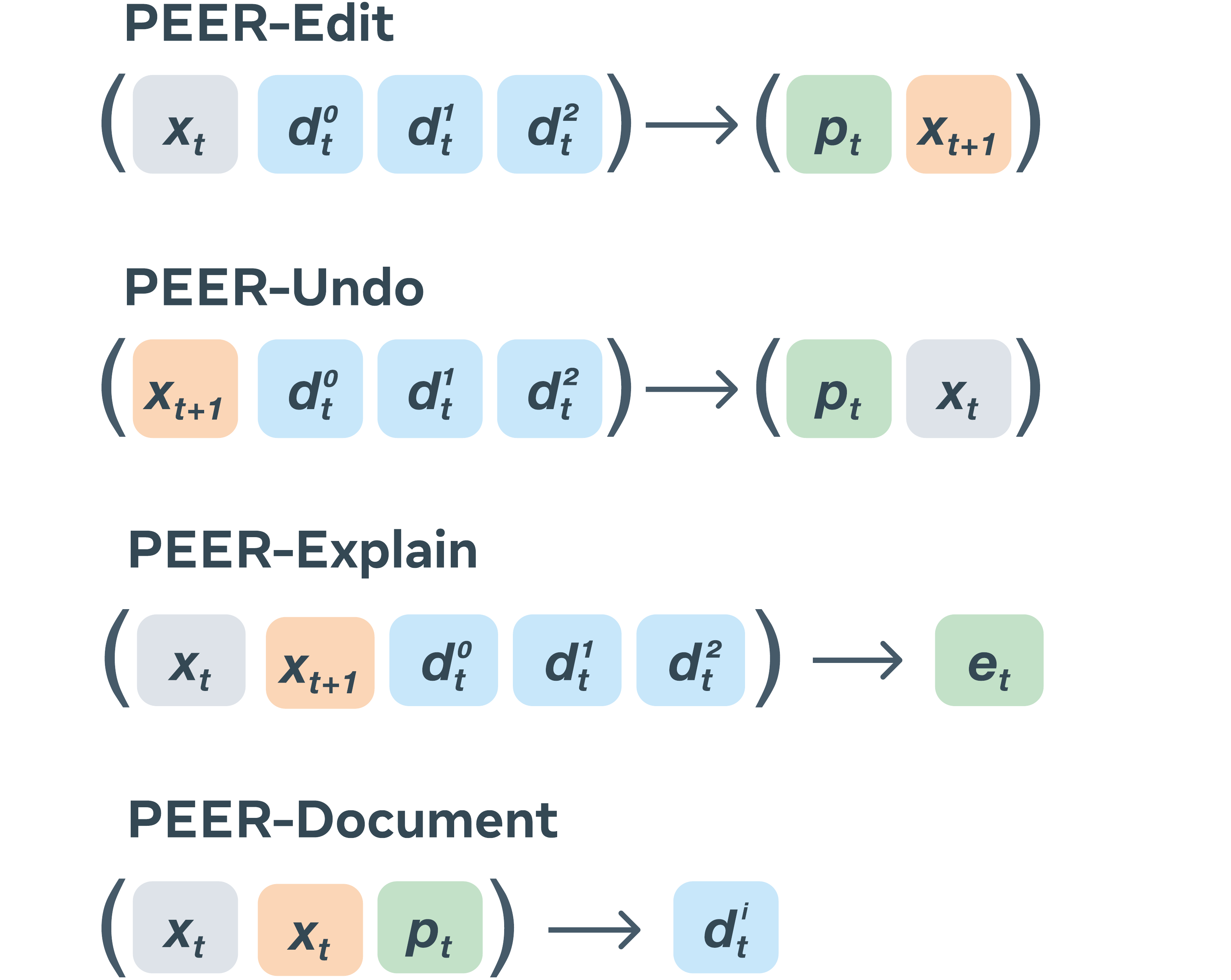}  
  \caption{To generalize to domains without editing histories, overcome data scarcity and improve the model's core abilities, we train various instances of PEER that perform different infilling tasks derived from this process.}
  \label{fig:peer_process_b}
\end{subfigure}
\caption{Schematic representation of the PEER process}
\label{fig:peer_process}
\end{figure*}

The core idea of our proposed framework is to model the editing of textual content as an \emph{iterative process}, where we repeatedly \emph{plan} and \emph{realize} changes (see Figure~\ref{fig:peer_process}a). Each iteration within this framework edits a text sequence $\mathbf{x}_t$ to obtain an updated version $\mathbf{x}_{t+1}$. For this edit, we assume that we are given a set of documents $D_t = \{d_t^1, \ldots, d_t^k\}$ containing relevant background information.\footnote{This set of documents aims to mirror the result of background research that humans often conduct before writing or editing factual texts. However, modeling this research itself is beyond the scope of this work, so we consider $D_t$ as given.} Given $\mathbf{x}_t$ and $D_t$, we first formulate a \emph{plan} $\mathbf{p}_t$---a rough idea of how the text should be modified, verbalized as a short text sequence like ``add more information'', ``fix grammar errors'' or ``use simpler language''. This plan is then realized by means of an actual \emph{edit} that transforms $\mathbf{x}_t$ into the updated state $\mathbf{x}_{t+1}$. Finally, the intention behind this edit can optionally be clarified by providing a textual \emph{explanation} $\mathbf{e}_t$; this is especially relevant in collaborative settings where explanations can facilitate evaluating the quality and usefulness of an edit \citep{liu-etal-2019-towards-explainable}. Note that the explanation can be similar or even identical to the plan, the conceptual difference being that the plan is made \emph{before} performing the edit, whereas the explanation is only formulated \emph{after} it was performed. 

The entire process of formulating a plan, collecting documents, performing an edit and explaining it, can be repeated multiple times to obtain a sequence of texts $\mathbf{x}_{t}, \mathbf{x}_{t+1}, \mathbf{x}_{t+2}, \ldots$ until either we arrive at some $\mathbf{x}_n$ for which $\mathbf{x}_n = \mathbf{x}_{n-1}$, or we reach a manually defined halting criterion. We can also write texts from scratch by starting with an empty sequence, i.e., $\mathbf{x}_0 = \varepsilon$. In reference to its four main parts, we refer to models based on this iterative process as PEER models.

\subsection{Overcoming Data Scarcity}
\label{sec:overcoming}

While using PEER to break the complex task of writing a coherent, consistent and factual document into many smaller subtasks has some potential benefits over standard left-to-right language modeling---such as being more interpretable and easier to control---it is challenging to find data from which this process can be learned at the scale required to train large language models. This is mainly because edit histories are difficult to obtain from web crawls, the most important data source for current language models \citep{brown2020language,rae2021gopher}. But even in cases where edit histories can be obtained (e.g., by collecting crawls of identical pages at different times) or synthetically generated, edits are typically not annotated with plans, documents, or explanations. 

Similar to prior work on text editing \citep{faltings-etal-2021-text,reid2022learning}, our first step in overcoming this issue is turning to Wikipedia -- a single source that comes close to fulfilling all our needs: It provides a full edit history including comments on a diverse set of topics, is large in scale, and articles frequently contain citations, which can be helpful for finding relevant documents. 
However, relying on Wikipedia as our sole source of training data comes with various severe downsides: First, it makes trained models specific to Wikipedia in terms of how they expect textual content to look like and what plans and edits they predict. Beyond that, comments in Wikipedia are noisy, and so in many cases, they are not an appropriate proxy for plans or explanations. Finally, numerous paragraphs in Wikipedia do not contain any citations; while this lack of background information can often be compensated by using a retrieval system \citep{piktus2021web,petroni-etal-2022-improving}, even such systems may not be able to find supporting background information for many edits.

\section{Infilling Edit Histories with PEER}

We propose a simple approach to address all issues that arise from Wikipedia being our only source of commented edit histories at once: We train not just one, but multiple instances of PEER that learn to infill various parts of the editing process (Figure~\ref{fig:peer_process}b); these models can then be used to generate synthetic data as a substitute for the missing pieces in our training corpus. In concrete terms, we train the following encoder-decoder models:

\begin{itemize}
    \item \textbf{\peerEdit{}}: Given an input text and a set of documents, this model learns to both plan and realize edits, i.e., it maps $(\mathbf{x}_t, D_t)$ to the sequence $(\mathbf{p}_t, \mathbf{x}_{t+1})$. This is done in an autoregressive fashion by factoring
    \begin{multline*}
    p(\mathbf{p}_t, \mathbf{x}_{t+1} \mid \mathbf{x}_t, D_t) \\ = \prod_{i=1}^n p(z_i \mid \mathbf{x}_t, D_t, z_1, \ldots, z_{i-1})
    \end{multline*}
    where $\mathbf{z} = z_1,\ldots,z_n = \mathbf{p}_t \cdot \mathbf{x}_{t+1}$ is the concatenation of $\mathbf{p}_t$ and $\mathbf{x}_{t+1}$.
    Thus, \peerEdit{} can update texts autonomously by generating both plans and edits, but it can also be provided with human-written plans as prefixes. As \peerEdit{} is our main model for actual editing, we also refer to it simply as PEER.
    
    \item \textbf{\peerUndo{}}: Given a text sequence $\mathbf{x}_{t+1}$ and a collection of documents $D_t$ that may have been used to write it, this PEER instance is trained to guess and \emph{undo} the latest edit by predicting the sequence $(\mathbf{p}_t, \mathbf{x}_t)$. This is done autoregressively analogous to \peerEdit{}.
    
    \item \textbf{\peerExplain{}}: This model is trained to autoregressively generate explanations $\mathbf{e}_i$ given $(\mathbf{x}_t, \mathbf{x}_{t+1}, D_t)$, i.e., an edit and a collection of relevant documents.
    
    \item \textbf{\peerDocument{}}: Given $(\mathbf{x}_t, \mathbf{x}_{t+1}, \mathbf{p}_t)$, this model is trained to generate a document $d \in D_t$ that provides useful background information for the edit.
    %\timo{In our last PEER meeting, Fabio mentioned that we might want to drop \peerDocument{} given that we only make use of it to improve a single skill (quoting a part of a cited passage) and it makes the method more convoluted. I'm unsure about this so I'm happy to get some more opinions :)}
\end{itemize}

\subsection{Synthetic Data Generation}

We use all variants of PEER to produce synthetic data---both to generate the missing pieces for completing our training data, and to \emph{replace} low-quality pieces in our existing data.

\paragraph{Decomposing Texts into Edits} To enable training on arbitrary text data even if it comes without edit histories, we use \peerUndo{} for generating synthetic ``backward'' edits: Given a plain text $\mathbf{x}\,{=}\,\mathbf{x}_n$ and a collection of documents $D$, we iteratively apply \peerUndo{} to obtain a sequence $(\mathbf{p}_{n-1}, \mathbf{x}_{n-1})$, $(\mathbf{p}_{n-2}, \mathbf{x}_{n-2}), \ldots$ until we arrive at some $\mathbf{x}_m = \varepsilon$. We can then train \peerEdit{} in the opposite direction, i.e., to predict each $(\mathbf{p}_t, \mathbf{x}_{t+1})$ from $\mathbf{x}_t$ and $D$, allowing us to use data from domains other than Wikipedia. 

\paragraph{Generating Plans} We use \peerExplain{} to address both the low quality of many comments in our corpus, and the fact that many edits simply do not have any comments. Given $\mathbf{x}_t$, $\mathbf{x}_{t+1}$ and a collection of documents $D_t$, we sample various outputs $\mathbf{e}_t^1, \ldots, \mathbf{e}_t^k$ from $\text{\peerExplain{}}(\mathbf{x}_t,\mathbf{x}_{t+1},D_t)$ that explain the edit being made and act as potential plans. We then compute the likelihood of the actual edit given each $\mathbf{e}_t^j$ and pick the one that makes this edit the most likely as its new plan:
\[
\hat{\mathbf{p}_t} = \argmax_{j \in \{1, \ldots,  k\}} p(\mathbf{x}_{t+1} \mid \mathbf{x}_t, D_t, \mathbf{e}_t^j)
\]
where $p(\mathbf{x}_{t+1} \mid \mathbf{x}_t, D_t, \mathbf{e}_t^j)$ denotes the probability that \peerEdit{} assigns to $\mathbf{x}_{t+1}$ given $\mathbf{x}_t$, $D_t$ and $\mathbf{e}_t^j$ as a plan.

\paragraph{Generating Documents} Finally, if we are unable to find relevant documents for a particular edit, we can use \peerDocument{} to generate synthetic documents containing information required to perform this edit. Crucially, we only do so for \emph{training} PEER-Edit; we never provide a model with any synthetic documents during inference. For generating documents, we proceed exactly as we do for generating new plans: We first sample a few documents from \peerDocument{} and then pick the one which helps \peerEdit{} the most in predicting the actual edit.

\subsection{Controlling Model Outputs}
\label{sec:controlling}

To improve the quality and diversity of generated plans, edits and documents, we implement control mechanisms similar to \citet{keskar2019ctrl} and \citet{he2020ctrlsum}---that is, we prepend specific \emph{control tokens} to the output sequences that a model is trained to generate, and then use these control tokens during inference to guide the model's generations. In particular, we make use of the following controls for different PEER models (see Appendix~\ref{appendix:controlling} for further details): 

\begin{itemize}
    \item For \peerExplain{}, we control the output \emph{length} as a proxy for the level of detail in generated explanations. We also control whether the generated comment starts with a verb in infinitive form; this approximates the notion of an \emph{instruction}, the format we expect humans to commonly use for communicating with PEER. Finally, we control whether there is a \emph{word overlap} between the explanation and the edit; preventing this during inference makes sure that generated plans do not make editing trivial by exactly specifying which words to add, remove or replace.
    \item For \peerUndo{}, we control the difference in the \emph{number of words} between $\mathbf{x}_{t+1}$ and $\mathbf{x}_t$. Through this, we can ensure that the sequence $\mathbf{x}_n, \mathbf{x}_{n-1}, \ldots$ eventually terminates at $\mathbf{x}_m = \varepsilon$ and does not get stuck in an infinite loop.
    \item For \peerDocument{}, we control whether the generated document \emph{contains a given substring}. This is useful when we want the document to contain a specific quote that is referred to in a Wikipedia edit.
\end{itemize}
We do not use any controls for \peerEdit{}, because---unlike for other models, which have specific and clearly defined tasks to solve---we do not make assumptions in advance about the types of editing tasks that users might want to solve with PEER-Edit and the kinds of control tokens that might be useful for these tasks.

\section{Training Data}
\label{sec:trainingdata}

Our main training data for PEER is derived from Wikipedia's edit history,\footnote{We use the February 2022 dump available at \url{https://dumps.wikimedia.org/enwiki/}.} which directly gives us access to raw tuples of source and target texts $(\mathbf{z}_t, \mathbf{z}_{t+1})$, from which we derive $\mathbf{x}_t$ and $\mathbf{x}_{t+1}$ after some preprocessing steps discussed below. Beyond that, the edit history also provides us with comments $\mathbf{c}_t$ that we use as proxies both for the plan $\mathbf{p}_t$ and for the explanation $\mathbf{e}_t$. Finally, as Wikipedia articles frequently use citations to back up claims, we can obtain an initial set $I_t$ of \emph{document identifiers} (e.g., URLs) for all documents cited in either $\mathbf{z}_t$ or $\mathbf{z}_{t+1}$. Our pipeline for transforming this raw data into the PEER format consists of three steps: First, we use some heuristics for filtering the data to remove low-quality edits and avoid overlap with any of our evaluation sets. We then use $I_t$ and a retrieval engine \citep{petroni-etal-2022-improving} to obtain a collection of relevant documents $D_t$ for each edit; finally, we convert the data into a format suitable for sequence-to-sequence models. In the following, we discuss all three preprocessing steps in more detail.

\paragraph{Filtering} As Wikipedia's edit history contains several low quality edits and numerous instances of vandalism \citep{potthast2008automatic}, we use some simple heuristics to improve data quality. In particular, we filter out edits that were reverted at some point and edits that were automatically made by bots without human involvement. Beyond that, we filter edits that affect more than two paragraphs and remove all edits for pages that are used in any of the datasets we evaluate on. We also discard examples where $I_t$ contains document identifiers that we are unable to resolve (e.g., because they refer to web pages that no longer exist). More details and a comprehensive overview of all filtering rules applied can be found in Appendix~\ref{appendix:filtering}.

\paragraph{Retrieving Documents} A crucial aspect of PEER is its use of documents that contain relevant background information. We thus aim to collect a set of documents $D_t = \{d_t^1, \ldots, d_t^k \}$ for each edit, with $k$ being a hyperparameter defining the number of documents. If $|I_t| \geq k$, we exclusively obtain these documents from citations that can be found in the edited paragraphs (i.e., in either $\mathbf{x}_t$ or $\mathbf{x}_{t+1}$). Otherwise, we augment these documents with ones from the Sphere corpus \citep{piktus2021web}, obtained using the pipeline described in \citep{petroni-etal-2022-improving}; further details are discussed in Appendix~\ref{appendix:retrieving}.

\paragraph{Formatting} Our first formatting step is to remove all paragraphs from $\mathbf{z}_t$ and $\mathbf{z}_{t+1}$ that are not affected by the edit. We then remove Wikipedia-specific syntax, but with a few exceptions:
We keep the syntax for representing titles, bold text, text in italics and lists, enabling the model to learn how to perform some basic formatting.
We also keep links and, more importantly, citations, enabling PEER to learn how to cite and quote from documents in $D_t$ to back up the textual content it generates. We denote the resulting text sequences with $\mathbf{z}'_t$ and $\mathbf{z}'_{t+1}$.

We linearize each document $d_t^i \in D_t$ using its content $\mathbf{c}_i$ and, if present, its title $\mathbf{t}_i$ and the corresponding web site's domain $\mathbf{d}_i$ as follows:
\[
\texttt{[}i\texttt{] } \mathbf{d}_i \texttt{ \# } \mathbf{t}_i \texttt{ \# } \mathbf{c}_i
\]
We include the number $i$ in this representation to facilitate citing and quoting specific documents. To finally obtain $\mathbf{x}_t$ and $\mathbf{x}_{t+1}$ from $\mathbf{z}'_t$ and $\mathbf{z}'_{t+1}$, we replace each citation of a document $d_t^i$ in both sequences with either
\[
\texttt{[[[} i \texttt{]]]} \text{\quad or\quad} \texttt{[[[} i \texttt{ quote=} \mathbf{q}_i \texttt{]]]}
\]
depending on whether in the original data, a specific subsequence $\mathbf{q}_i$ of $d_t^i$ was quoted.
As $\mathbf{p}_t$ and $\mathbf{e}_t$ are already simple text sequences, we do not perform any modifications to them.

If there are multiple inputs or outputs, we simply concatenate them using a special separator sequence. Moreover, if the text we are editing has a title, we always prepend this title to the original input sequence. An example of a linearized input for PEER is illustrated in Figure~\ref{fig:peer_linearized}.

\section{Experiments}

We conduct a series of experiments to investigate whether---despite Wikipedia being our only \emph{natural} source of comments and edits---our infilling techniques enable us to turn PEER into a general purpose editing model that is capable of following human-written plans and tackling a wide range of collaborative editing tasks in different domains at satisfactory performance. More specifically, we aim to answer the following research questions:

\begin{itemize}
    \item Can PEER follow plans and perform meaningful edits in domains for which no edit histories are available, and does our self-training approach using \peerUndo{} to generate synthetic edits improve this ability? (Section~\ref{sec:naturaledits})
    \item Does the ability to follow plans based on Wikipedia comments transfer to instructions specified by humans, and can it be improved by training on synthetic plans generated using \peerExplain{}? (Section~\ref{sec:downstream})
    \item Can PEER make proper use of citations and quotes to explain generated outputs, and can \peerDocument{} amplify this? (Section~\ref{sec:citations})
    \item How does writing text sequences in a single pass compare to an iterative application of PEER, both if the model runs autonomously and if it is provided with human-written plans? (Section~\ref{sec:language})
\end{itemize}
Experiments conducted to answer these research questions are complemented by a qualitative analysis of model outputs and exemplary sessions of human-AI interactions in Section~\ref{sec:analysis}.

\subsection{Experimental Setup}

We initialize all instances of PEER from \emph{LM-Adapted} T5 \citep{raffel2019exploring}; by default, we use the variant with 3B parameters. Each model is trained for 20,000 steps on 64 GPUs with an effective batch size of 256, corresponding to about five million Wikipedia edits. The maximum sequence length is set to 1,024 and 384 tokens for input and output, respectively. We set $k = 3$ as the maximum number of reference documents per example.

We use a variety of metrics to evaluate PEER and our baseline models on all tasks considered:

\begin{itemize}
    \item \textbf{Exact Match} (EM) is the percentage of examples for which the performed edit exactly matches a given target;
    \item \textbf{EM-Diff} is a variant of EM that is computed on the diff level;\footnote{Diffs are obtained using Python's \texttt{difflib} library. For a model output $\mathbf{x}_{t+1}'$, we compute EM-Diff as $|\text{diff}(\mathbf{x}_t, \mathbf{x}_{t+1}) \cap \text{diff}(\mathbf{x}_t, \mathbf{x}'_{t+1}) | \div \max(|\text{diff}(\mathbf{x}_t, \mathbf{x}_{t+1})|, |\text{diff}(\mathbf{x}_t, \mathbf{x}'_{t+1})|)$.}
    \item \textbf{SARI} \citep{xu-etal-2016-optimizing} averages match scores for the three word-level edit operations \emph{add}, \emph{delete} and \emph{keep};\footnote{We use the SARI implementation of EASSE \citep{alva-manchego-etal-2019-easse}.}
    \item \textbf{GLEU} \citep{napoles-etal-2015-ground} is a variant of BLEU \citep{papineni-etal-2002-bleu} proposed for grammatical error correction tasks;
    \item \textbf{Rouge} \citep{lin2004rouge} is a set of metrics based on $n$-gram overlap (Rouge-$n$) or longest common subsequences (Rouge-L); 
    \item \textbf{Update-Rouge} \citep{logan2021fruit} is a variant of Rouge that is computed only on sentences updated during an edit.
\end{itemize}

\subsection{Natural Edits}
\label{sec:naturaledits}

We first evaluate PEER's ability to follow a diverse set of plans, leverage provided documents and perform edits across different domains; in particular, we are interested in investigating its performance in domains for which no edit histories are available. To this end, we introduce \emph{Natural Edits}, a collection of naturally occuring edits for different text types and domains that we obtain from three English web sources: We collect encyclopedic pages from Wikipedia, news articles from Wikinews, and questions from the \emph{Cooking}, \emph{Gardening}, \emph{Law}, \emph{Movies}, \emph{Politics}, \emph{Travel} and \emph{Workplace} subforums of StackExchange. All of these sites provide edit histories with comments that often elaborate on the edit's intent and that we provide to all models as plans.\footnote{The ability of PEER to follow plans based on human-written instructions is investigated in Section~\ref{sec:downstream}.} We split each dataset into training and test data. However, we only provide plain texts instead of actual edits in the training sets of the Wikinews and StackExchange subsets, enabling us to test editing abilities in domains for which no edit histories are accessible. Relevant  statistics for Natural Edits are shown in Table~\ref{tab:natural_edits}.

\begin{table}
    \centering
    \small
    \begin{tabularx}{\linewidth}{Xrrrc}
    \toprule
    \textbf{Subset} & \textbf{Train (Edit)} & \textbf{Train (PT)} & \textbf{Test}  & \textbf{Doc.}  \\
    \midrule
    Wikipedia & 6,960,935 & -- & 4,000 & \checkmark \\
    Wikinews & -- & 125,664 & 1,000 & -- \\
    Cooking & -- & 22,517 & 500 & -- \\
    Gardening & -- & 13,258 & 500  & -- \\
    Law & -- & 16,418 & 500 & -- \\
    Movies & -- & 19,601 & 500 & -- \\
    Politics & -- & 10,676 & 500 & -- \\
    Travel & -- & 38,961 & 500 & -- \\
    Workplace & -- & 18,231 & 500 & -- \\
    \bottomrule
    \end{tabularx}
    \caption{Overview of the number of edits and plain texts (PT) in the train sets and the number of edits in the test sets of Natural Edits. The final column shows whether the subset uses reference documents.}
    \label{tab:natural_edits}
\end{table}

As a first experiment, we check whether PEER actually learns to make use of provided documents and plans by evaluating it on the Wikipedia subset of Natural Edits. We compare regular PEER provided with gold plans to variants trained and evaluated (i) without plans, (ii) without reference documents, and (iii) without both plans and reference documents. Table~\ref{tab:ne_wiki} shows EM, EM-Diff and SARI scores for all models and a copying baseline, for which $\mathbf{x}_{t+1} = \mathbf{x}_t$. As can be seen, PEER substantially outperforms all baselines. PEER without both plans and documents performs much worse than just removing one of both, illustrating that plans and documents provide complementary information that the model is capable of using; this is in line with findings of \citet{faltings-etal-2021-text}. 
%Plans are more important in terms of EM and EM-Diff, which are most influenced by simple edits such as fixing grammar errors. On the other hand, documents are more useful in terms of SARI, which is more meaningful for substantial edits such as adding entirely new sentences. 
% PEER and \peerEditSynPlans{} perform similarly, showing that training on synthetic plans has little impact on performance if the plans used during evaluation are from the same low-quality distribution as the original plans. Finally, Table~\ref{tab:ne_wiki} also shows results for an 11B parameter variant of \peerEditSynPlans{}, which achieves the best results overall.

\begin{table}
    \centering
    \small
    \begin{tabularx}{\linewidth}{Xccc}
    \toprule
    \textbf{Model} & \textbf{EM} & \textbf{EM-Diff} & \textbf{SARI}  \\
    \midrule
    Copy    & \phantom{0}0.4 & \phantom{0}0.0 & 32.7 \\
    PEER    & \textbf{23.1} & \textbf{26.2} & \textbf{55.5} \\
    %\peerEditSynPlans{} & 22.9 & \textbf{26.2} & \textbf{57.0} \\
    PEER (no plans) & 18.0 & 19.8 & 52.0 \\
    PEER (no documents)  & 19.8 & 22.8 & 51.7 \\
    PEER (no plans/documents) & 13.5 & 15.1 & 45.9 \\
    %\midrule
    %\peerEditSynPlans{} (11B) & \textit{25.5} & \textit{29.0} & \textit{58.0} \\
    % \peerEditSynPlans{} (11B + more data) & \qm & \qm & \qm \\
    \bottomrule
    \end{tabularx}
    \caption{Results for variants of PEER on the Wikipedia subset of Natural Edits. Plans and documents provide complementary information and substantially improve performance.}
    \label{tab:ne_wiki}
\end{table}

\begin{table*}
    \centering
    \setlength{\tabcolsep}{2.7pt}
    \small
    \begin{tabularx}{\linewidth}{Xccccccccc}
    \toprule
                & \textbf{Wiki} & \textbf{News} & \textbf{Cooking} & \textbf{Garden} & \textbf{Law} & \textbf{Movies} & \textbf{Politics} & \textbf{Travel} & \textbf{Workpl.} \\
    \midrule
Copy            & \phantom{0}0.0 / 32.7 & \phantom{0}0.1 / 32.8 & \phantom{0}0.0 / 31.6 & \phantom{0}0.0 / 32.0 & \phantom{0}0.0 / 31.1 & \phantom{0}0.0 / 31.5 & \phantom{0}0.0 / 31.8 & \phantom{0}0.0 / 31.2 & \phantom{0}0.0 / 31.5 \\
PEER\,(no\,plans) & 16.6 / 50.7 & 10.8 / 41.3 & \phantom{0}4.5 / 36.3 & \phantom{0}1.8 / 35.1 & \phantom{0}2.6 / 35.8 & \phantom{0}2.9 / 35.3 & \phantom{0}2.1 / 36.5 & \phantom{0}1.6 / 34.8 & \phantom{0}3.1 / 34.7 \\
PEER            & \textbf{26.2} / \textbf{55.5} & 21.3 / 49.3 & 11.0 / 40.2 & \phantom{0}4.4 / 37.7 & \phantom{0}7.5 / 36.4 & \phantom{0}6.7 / 39.2 & \phantom{0}6.8 / 38.7 & \phantom{0}6.7 / 38.1 & \phantom{0}6.9 / 36.7 \\
PEER (DA)    & -- &  \textbf{23.3} / \textbf{51.6} & \textbf{13.2} / \textbf{42.9} & \textbf{\phantom{0}8.1} / \textbf{44.9} & \textbf{\phantom{0}9.4} / \textbf{39.0} & \textbf{\phantom{0}9.9} / \textbf{42.4} & \textbf{11.6} / \textbf{41.3} & \textbf{\phantom{0}9.1} / \textbf{40.2} & \textbf{\phantom{0}8.3} / \textbf{39.2} \\

    \bottomrule
    \end{tabularx}
    \caption{EM-Diff / SARI scores on all subsets of Natural Edits. The domain-adapted (DA) variants of PEER clearly outperform regular PEER, demonstrating the usefulness of synthetic edits generated with \peerUndo{}.}
    \label{tab:ne_all}
\end{table*}

Next, we evaluate PEER on all subsets of Natural Edits in order to assess its ability to perform edits in different domains. We use \peerUndo{} as described in Section~\ref{sec:overcoming} to create synthetic edits from plain texts and train domain-adapted (DA) variants of PEER. For generating synthetic edits, we found it sufficient to apply \peerUndo{} just once for each plain text $\mathbf{x}_t$ to obtain a tuple $(\mathbf{p}_{t-1}, \mathbf{x}_{t-1})$. Upon manual inspection, we also found that the generated plans $\mathbf{p}_{t-1}$ do not actually match the undone edit, so we use \peerExplain{} as described in Section~\ref{sec:overcoming} to rewrite all plans. We finetune the domain-adapted variants of PEER-Edit on a balanced mixture of examples from the original training distribution and synthetic in-domain edits for 1,000 steps; we do so separately for the Wikinews and StackExchange subsets of Natural Edits, resulting in two instances of domain-adapted PEER. Results on Natural Edits are shown in Table~\ref{tab:ne_all}, which reports both EM-Diff and SARI scores across all subsets. As can be seen, plans are extremely helpful across domains, indicating that the ability to understand plans found in Wikipedia edits directly transfers to other domains. Importantly, the domain-adapted variants of PEER clearly outperform regular PEER for all subsets of Natural Edits, with particularly strong improvements on the Gardening, Politics, and Movies subsets (84\%, 71\% and 48\% EM-Diff, respectively). This demonstrates the effectiveness of generating synthetic edits for applying PEER in different domains.

\subsection{Downstream Tasks}
\label{sec:downstream}

So far, we have only evaluated PEER using plans based on naturally occurring \emph{comments}. But to what extend is it capable of following \emph{instructions} formulated by humans to yield well known editing functionalities, and can training on synthetic plans improve this ability? To answer these questions, we next evaluate PEER on various downstream editing tasks in a zero-shot fashion. For this evaluation, we consider the following datasets:

\begin{itemize}
    \item \textbf{JFLEG} \cite{napoles2017jfleg} is a grammatical error correction dataset with single-sentence inputs written by English language learners;
    \item \textbf{ASSET} \cite{alva2020asset} is a crowdsourced corpus for single-sentence text simplification;
    \item \textbf{\textsc{IteraTeR}} \cite{du2022understanding} is an editing dataset spanning five edit intentions across three different domains;\footnote{We only include edits from the \emph{non-meaning-changed} categories ``fluency'', ``coherence'' and ``clarity'' in our evaluation.}
    \item \textbf{WNC} \cite{pryzant2020automatically} is a dataset where the task is to remove or mitigate biased words to make sentences more neutral;
    \item \textbf{FRUIT} \cite{logan2021fruit} contains texts from Wikipedia that need to be \emph{updated}; for performing this update, various reference documents from Wikipedia are provided;
    \item \textbf{\textsc{WAFER-Ins}} is based on the WAFER dataset~\citep{petroni-etal-2022-improving}; the task is to \emph{insert} a sentence at some position in a Wikipedia paragraph given documents from the Sphere corpus~\citep{piktus2021web} that contain relevant background information.
\end{itemize}

In addition to PEER-Edit, we also consider a variant trained with synthetic plans; that is, we replace each original plan with one generated by \peerExplain{} as described in Section~\ref{sec:overcoming}. We refer to the \peerEdit{} variant trained on these synthetic plans as \peerEditSynPlans{}. When generating synthetic plans, we use the control tokens introduced in Section~\ref{sec:controlling} to ensure a diverse set of plan lengths. For 80\% of generated plans, we enforce that they start with a verb and have no word overlap with the performed edit, respectively. Details are discussed in Appendix~\ref{appendix:controlling}.

We compare PEER and \peerEditSynPlans{} to various baseline models that were either trained in a fully unsupervised fashion, or, in a similar spirit as PEER, were trained to solve tasks given human-written instructions; however, these instructions are not \emph{naturally occurring} like the Wikipedia comments we use for PEER, but explicitly written by humans to create training datasets for finetuning language models. While this generally results in higher quality data, such datasets are also much more difficult and expensive to obtain. Thus, if we are able to bridge the gap between noisy comments and actual instructions, PEER can provide a much less expensive alternative. In concrete terms, we compare PEER to the following models:

\begin{itemize}
    \item \textbf{T$k$-Instruct} \citep{wang2022benchmarking} is, like PEER, initialized from the \emph{LM Adapt} variant of T5. It is finetuned on Natural Instructions v2, a collection of instructions for more than 1,600 tasks, including grammatical error correction and text simplification.
    \item \textbf{T0} and \textbf{T0++} \citep{sanh2022multitask} are also initialized from the \emph{LM Adapt} variant of T5; they are then finetuned using a variety of human-written prompts from \emph{PromptSource} \citep{bach-etal-2022-promptsource}. Unlike Natural Instructions v2, this dataset does not directly contain editing tasks, but related tasks including summarization and data-to-text generation.
    \item \textbf{GPT3} \citep{brown2020language} is a pretrained decoder-only model that is not finetuned on any instructions; with 175B parameters, it is larger than our default PEER models by two orders of magnitude. We also compare to \textbf{InstructGPT} \citep{ouyang2022training}, a variant of GPT3 that was finetuned on a large dataset of instructions and corresponding outputs written by humans; neither the dataset nor the model parameters are publicly available, so we access the models via OpenAI's API.\footnote{We use the \emph{text-davinci-001} variant described in \citep{ouyang2022training}. The OpenAI API also provides a more recent version, \emph{text-davinci-002}, but as of this writing, we could not find any publicly available information about the data used to train this version or other training details.}
    \item \textbf{OPT} \citep{zhang2022opt} is an open-source replica of GPT3; it is not finetuned on any labeled data.
\end{itemize}

As all of these models are capable of processing textual prompts, we formulate a single plan $\mathbf{p}$ per task that we provide to all models. However, we embed this plan into slightly different contexts to make it most suitable for each model (for example, T$k$-Instruct expects the plan to be prefixed by the string ``Definition:''). For FRUIT and \textsc{WAFER-Ins}, we also use a more natural format than the one used by PEER for providing references to our baseline models; all prompts and model-specific modifications are shown in Appendix~\ref{appendix:prompts}. 

Unless otherwise specified, we use greedy decoding for all models. We do not perform any task-specific finetuning or in-context learning as we are interested in evaluating each model's suitability as a \emph{general} editing model: In the general case of a user providing a plan, we cannot assume access to other examples using the exact same plan. Beyond that, especially for tasks that require references, we typically cannot fit more than one example into the context window.
We do not compare to other editing models trained on Wikipedia data, as they are either only trained to solve specific tasks \citep{logan2021fruit}, not able to follow instructions \citep{reid2022learning} or only capable of processing single-sentence inputs \citep{faltings-etal-2021-text}; beyond that, none of these models are publicly available. However, we additionally report supervised state-of-the-art scores for all tasks considered.

\begin{table*}
\newcolumntype{Y}{>{\centering\arraybackslash}X}
    \centering
    \small
    \setlength{\tabcolsep}{3pt}
    \begin{tabularx}{\linewidth}{llr@{\hskip 0.7cm}YccY@{\hskip 0.7cm}Yc@{\hskip 0.7cm}c}
    \toprule
    & & & \multicolumn{4}{c@{\hskip 0.7cm}}{\textbf{Without Documents}} & \multicolumn{2}{c@{\hskip 0.7cm}}{\textbf{With Documents}} & \\
    & \textbf{Model} & \textbf{Params} & JFLEG & ASSET & \textsc{IteraTeR} & WNC & FRUIT & WAFER\textsc{-Ins} & \textbf{Avg} \\
    \midrule
    \multirow{7}{*}{\textbf{(a)}} & Copy & \multicolumn{1}{c}{--} & 26.7 / 40.5  & 20.7 & 30.5 & 31.9 / \phantom{0}0.0 & 29.8 / \phantom{0}0.0 & 33.6 / --\negphantom{--}\phantom{00.0} & 28.9 \\ 
    & T$k$-Instruct & 3B & 31.7 / 38.7 & 28.3 & 36.2 & 30.3 / \phantom{0}0.0 & 12.7 / \phantom{0}3.9 & \phantom{0}1.6 / --\negphantom{--}\phantom{00.0} & 23.5 \\
    & T0 & 3B & 42.9 / 38.6 & 28.6 & 28.1 & 17.8 / \phantom{0}0.0 & 13.1 / \phantom{0}5.7 & \phantom{0}6.1 / --\negphantom{--}\phantom{00.0} & 22.8 \\
    & T0++ & 11B & 35.9 / 43.8 & 25.8 & 36.1 & 27.0 / \phantom{0}0.0 & 16.1 / \phantom{0}3.7 & \phantom{0}3.9 / --\negphantom{--}\phantom{00.0} & 24.1 \\ 
    & PEER & 3B & 54.8 / 55.1 & 29.9 & 36.5 & 56.4 / 31.9 & 39.4 / 28.3 & 35.2 / 33.6 & 42.0 \\ 
    & \peerEditSynPlans{} & 3B & 59.0 / 57.2 & \textbf{33.2} & 37.1 & 56.6 / 32.7 & 40.3 / \textbf{33.9} & 35.5 / 37.6 & 43.6 \\
    & \peerEditSynPlans{} & 11B & \textbf{59.9} / \textbf{58.6} & 32.4 & \textbf{37.8} & \underline{\textbf{58.8}} / \underline{\textbf{34.7}} & \textbf{40.7} / 33.5 & \underline{\textbf{35.9}} / \underline{\textbf{38.4}} & \underline{\textbf{44.3}} \\ 
    \midrule
    \multirow{2}{*}{\textbf{(b)}} & \peerEditSynPlans[, i3]{} & 3B & \underline{63.3} / 59.6 & 36.1 & 37.1 & 45.2 / 12.4 & \underline{41.6} / 34.6 & 35.2 / 37.0 & 43.1 \\ 
    & \peerEditSynPlans[, s-i3]{} & 3B & 57.4 / 49.7 & \underline{40.7} & 35.8 & 38.4 / \phantom{0}3.9 & \underline{41.6} / \underline{38.7} & 32.9 / 34.3 & 41.1 \\ 
    \midrule
    \multirow{3}{*}{\textbf{(c)}} & OPT & 175B & 49.2 / 49.4 & 25.8 & 31.4 & 25.1 / \phantom{0}0.0 & 35.6 / 27.4 & 21.1 / --\negphantom{--}\phantom{00.0} & 31.4 \\
    & GPT3 & 175B & 50.6 / 51.8 & 25.0 & 30.7 & 26.0 / \phantom{0}0.5 & 33.6 / 25.9 & 22.9 / --\negphantom{--}\phantom{00.0} & 31.5 \\ 
    & InstructGPT & 175B & 62.3 / \underline{60.0} & 35.4 & \underline{38.2} & 33.9 / \phantom{0}0.7 & 37.5 / 23.4 & 29.2 / --\negphantom{--}\phantom{00.0} & 39.4 \\ 
    \midrule
    \multirow{1}{*}{\textbf{(d)}} & Sup. SotA & \multicolumn{1}{c}{--} & \negphantom{--}\phantom{00.0}-- / \textit{62.4} & \textit{44.2} &  \textit{37.2} & \negphantom{--}\phantom{00.0}-- / \textit{45.8} & \negphantom{--}\phantom{00.0}-- / \textit{47.4} & -- & -- \\ 
    \bottomrule
    \end{tabularx}
    \caption{Downstream task results for PEER and various baselines, divided into four groups: \textbf{(a)} T5-based models and a copy baseline, \textbf{(b)} PEER with different sampling strategies, \textbf{(c)} 175B parameter decoder-only models, \textbf{(d)} supervised state of the art. The first numbers for each task are SARI scores; additional metrics are GLEU for JFLEG, EM for WNC, Update-R1 for FRUIT and SARI scores obtained if the model is told exactly where to insert a new sentence for WAFER-\textsc{Ins}. Supervised scores from left to right are from \citet{ge2018reaching}, \citet{martin2020muss}, \citet{du2022understanding}, \citet{pryzant2020automatically} and \citet{logan2021fruit}, respectively. The best result for models based on \emph{LM Adapted} T5 is shown in bold, the best zero-shot performance overall is underlined. On average, \peerEditSynPlans{} clearly outperforms all baselines.}
    \label{tab:downstream_results}
\end{table*}

Results are shown in Table~\ref{tab:downstream_results}, with rows grouped into four different sets. The first group contains a copying baseline and all models based on \emph{LM Adapted} T5. PEER substantially outperforms all other models in this group, with the 3B model achieving an average SARI score of 42.0 across all tasks, compared to 24.1 for the strongest T5-based baseline model. Importantly, \peerEditSynPlans{} consistently outperforms regular PEER, increasing the average score by 1.6 points. This clearly demonstrates the usefulness of generating synthetic plans to enhance PEER's ability to follow instructions. Increasing the model size to 11B parameters slightly improves results for most tasks.

Given the iterative nature of PEER's editing process, the second group considers an alternative decoding strategy where PEER is applied multiple times; we consider both greedy decoding for three iterations (i3) and top-$p$ sampling \citep{Holtzman2020The} with $p = 0.9$ for three iterations (s-i3). As can be seen, different decoding strategies sometimes drastically improve results (e.g., for ASSET), but can also lead to much worse performance (e.g., for WNC). We thus believe that automatic ways of finding the best decoding strategy for a given plan is an interesting avenue for future work.

All 175B parameter models are shown in the third group. OPT and GPT3 perform much worse than PEER despite being much larger. InstructGPT comes close to and even outperforms PEER for some tasks that do not require documents; however, it clearly lags behind PEER when it comes to handling documents for updating text or adding new information. Averaged across all tasks, it performs 4.1 points worse than \peerEditSynPlans{} (3B), despite being much larger \emph{and} being finetuned on human-annotated data. We also note that, given a high-quality corpus with human annotations, PEER could easily be further finetuned in a similar fashion to InstructGPT.

Finally, the last row of Table~\ref{tab:downstream_results} shows supervised state-of-the-art performance. While our zero-shot models clearly lag behind these scores on average, they approach supervised performance in some cases.

\subsection{Citations and Quotes}
\label{sec:citations}

Unlike our baseline models, PEER is capable of both \emph{citing} and \emph{quoting} from reference documents to back up the claims it generates. This is useful both in terms of explainability and verifiability, as it allows users to easily fact-check these claims. The ability to quote from individual passages---as opposed to citing an entire document---is especially helpful for long documents, which can take some time to process entirely.

To facilitate the evaluation of PEER's ability to cite and quote, we consider both tasks in isolation. To this end, we introduce two new datasets based on Natural Edits: \emph{\textsc{Ne}-Cite} and \emph{\textsc{Ne}-Quote}. For building these datasets, we collect examples from Wikipedia's edit history where the only difference between $\mathbf{x}_t$ and $\mathbf{x}_{t+1}$ is that a new citation was added for \textsc{Ne}-Cite, or that a quote was added to an existing citation for \textsc{Ne}-Quote. Naturally, we make sure that the cited document is always present in the $k=3$ documents provided to PEER. To make the task of citing the correct document challenging, we obtain the other two documents in $D_t$ by applying the BM25 \citep{robertson1995okapi} and DPR \citep{karpukhin2020dense} variants of \citet{petroni-etal-2022-improving} to find the best match in Sphere \citep{piktus2021web}, respectively. If the gold document contains too many tokens, for \textsc{Ne}-Cite we pick the best chunk according to the reranker of \citet{petroni-etal-2022-improving}; for \textsc{Ne}-Quote, we select the chunk from the document that actually contains the quote. In total, we collect 2,351 and 391 examples, respectively, for which we manually set the plans to simply be ``Add a citation'' and ``Add a quote''.

Importantly, PEER's training data contains only very few examples of edits using quotes. This is mainly because quotes are used sparingly in Wikipedia. Moreover, we are unable to use the vast majority of examples containing quotes because they often come from non-online sources or web pages that no longer exist, so we do not have access to the documents that the quotes are taken from. To overcome this issue, we use \peerDocument{} to write synthetic documents for all edits that add quotes and for which the actual document is missing; this gives us an additional 8,739 training examples. We finetune PEER on these examples, mixed with around 500k examples from the original distribution, for 2,000 steps; we refer to this variant trained with synthetic quotes as PEER (SQ).

\begin{table}
    \centering
    \small
    \setlength{\tabcolsep}{1.9pt}
    \begin{tabularx}{\linewidth}{Xcc@{\hskip 0.2cm}c}
    \toprule
    \textbf{Model} & \textbf{\textsc{Ne}-Cite} & \textbf{\textsc{Ne}-Quote} & \textbf{\textsc{Ne}-Quote (con.)} \\
    \midrule
    Random & \negphantom{--}\phantom{00.0}-- / 33.3 & -- & 40.1 / 31.7 / 36.5 \\ 
    Unigram & \negphantom{--}\phantom{00.0}-- / 34.2 & -- & -- \\
    Side & \negphantom{--}\phantom{00.0}-- / \textbf{91.1} & -- & -- \\
    Lead & -- / -- & -- & 50.6 / 44.0 / 46.0 \\
    \midrule
    PEER & 74.1 / 88.1 & \phantom{0}0.0 / \phantom{0}0.0 / \phantom{0}0.0 & 49.3 / 44.3 / 48.1 \\ 
    \peerEditSynPlans & 74.5 / 88.9 & \phantom{0}0.2 / \phantom{0}0.1 / \phantom{0}0.1 & 49.8 / 44.8 / 48.7 \\ 
    PEER\,(SQ) & \textbf{74.9} / 87.9 & \textbf{13.6} / \textbf{11.9} / \textbf{12.9} & \textbf{58.1} / \textbf{54.6} / \textbf{57.3} \\ 
    \bottomrule
    \end{tabularx}
    \caption{Accuracy on \textsc{Ne}-Cite (without/with gold positions) and R1/R2/RL scores on both \textsc{Ne}-Quote and constrained \textsc{Ne}-Quote. When given the correct position, \peerEditSynPlans{} almost matches the performance of the supervised Side model on \textsc{Ne}-Cite, demonstrating its strong citing abilities. Training on synthetic documents substantially improves PEER's ability to quote relevant passages.}
    \label{tab:ne_cq}
\end{table}

For \textsc{Ne}-Cite, we use the percentage of times where the correct document was cited \emph{and} the citation was placed at the right position as our evaluation metric. We compare PEER's citing abilities to three baselines: randomly picking a reference, selecting the reference that maximizes the unigram overlap with $\mathbf{x}_t$, and using the \emph{Side} reranker \citep{petroni-etal-2022-improving}, a model trained on millions of actual Wikipedia citations. Noticeably, none of these baselines is able to decide \emph{where} to place the citation; besides, the Side reranker needs to be given the position of the citation to be added. We thus also consider a variant of \textsc{Ne}-Cite where models are told where to place the citation; for PEER, this is achieved using a decoder prefix that is identical to the input up to this position, at which it adds the sequence \texttt{[[[} to indicate the start of a citation. Scores both without and with providing the correct positions are shown in Table~\ref{tab:ne_cq}. If not provided with the correct position, PEER puts the right citation at the right place in 74.1\% of cases, with \peerEditSynPlans{} slightly improving performance. When given the correct position, \peerEditSynPlans{} even comes close to the supervised Side model (88.9 vs 91.1), clearly outperforming the other baselines. Finetuning on synthetic quotes does not significantly alter PEER's citing ability.

Similar to citing, we also look at two variants of the quoting task: In the first variant, the model needs to add a quote without being told where to put it and which document to quote from; in the second variant, the model is given access to the required position of the quote in a similar fashion to our approach for \textsc{Ne}-Cite (i.e., by adding a decoder prefix that ends with the string \texttt{quote=}). For this variant, we additionally use constrained decoding as proposed by \citet{cao2021autoregressive} to ensure that the generated quote is actually contained in the cited document. We compare to two baselines: One that selects a random substring of $n$ words from the gold document, and one that selects the lead $n$ words; we set $n$ to be the median length of quotes in \textsc{Ne}-Quote. As a metric, we report Rouge-1/2/L scores \citep{lin2004rouge} computed only on quotes, which we set to 0 in case a model does not add a quote at all. As shown in Table~\ref{tab:ne_cq}, PEER and \peerEditSynPlans{} are completely unable to quote without decoder prefixes and constrained decoding. Training on synthetic documents improves performance, but still results in rather low scores. Adding decoder prefixes and using constrained decoding improves performance a lot, but PEER still does not outperform the lead baseline. However, PEER (SQ) achieves much stronger results in this setting, improving R1/R2/RL scores by 7.5, 10.6 and 11.3 points over the lead baseline, respectively; this demonstrates the effectiveness of using \peerDocument{} to generate synthetic documents for improving \peerEdit's ability to explain generated claims by quoting from provided documents.

\subsection{Iterative Editing for Text Generation}
\label{sec:language}

Finally, we investigate PEER's ability to generate new texts from scratch, i.e., to perform a series of edits starting from $\mathbf{x}_0 = \varepsilon$. To this end, we collect a set of 500 intro sections from Wikipedia, each with three reference documents. As a baseline, we finetune \emph{LM Adapted} T5 as a conditional language model on the exact same data that PEER was trained on -- that is, the model is trained to predict $\mathbf{x}_{t+1}$ given $D_t$ and the page's title, but not $\mathbf{x}_t$. However, we use a special character sequence to inform the model about whether $\mathbf{x}_{t+1}$ is an intro section; we train this baseline, that we refer to as \emph{WikiLM}, with the exact same parameters as PEER.

We evaluate PEER in three different modes: 
\begin{itemize} 
\item an \emph{autonomous} mode, where the model continuously writes and realizes its own plans without human involvement;
\item a \emph{manual} mode, where we give the model a series of human-written plans that it is supposed to realize. We choose a simple sequence of three plans that we use for all intros: $\mathbf{p}_0 = $ ``Create page'', followed by $\mathbf{p}_1 = \mathbf{p}_2 =$ ``Add more information'';
\item a \emph{collaborative} mode, where human-written plans are interleaved with plans proposed by PEER; that is, we use the plan sequence $\mathbf{p}_0, \mathbf{p}_0', \mathbf{p}_1, \mathbf{p}_1', \mathbf{p}_2$, where $\mathbf{p}_0$, $\mathbf{p}_1$ and $\mathbf{p}_2$ are as above, whereas PEER generates $\mathbf{p}_0'$ and $\mathbf{p}_1'$.
\end{itemize}

Without controlling for output length, WikiLM generates rather short intros, resulting in relatively low Rouge-1/2/L scores. To make the comparison more fair, we thus split our dataset of Wikipedia intros into 100 dev examples and 400 test examples; the dev examples are exclusively used for picking the exponential \emph{length penalty} \citep{murray-chiang-2018-correcting} that maximizes the model's average Rouge-1 score. We also prevent models from generating the same token 5-gram more than once to avoid endless repetitions.

\begin{table}
    \centering
    \small
    \begin{tabularx}{\linewidth}{Xccc}
    \toprule
    Model & LP & R1 / R2 / RL & QE \\
    \midrule
    Wiki-LM     & 5.0 & 38.4 / 16.9 / 27.3 & 38.7 \\ 
    PEER (autonomous)       & 5.0 & 37.7 / 15.8 / 26.2 & 40.6 \\
    PEER (manual)  & 2.0 & 39.4 / 17.0 / 28.1 & \textbf{41.1} \\
    PEER (collaborative)   & 2.0 & \textbf{39.5} / \textbf{17.2} / \textbf{28.4} & 41.0  \\
    \bottomrule
    \end{tabularx}
    \caption{Results for various approaches on our Wikipedia intro generation test set. Length penalty (LP) is optimized on the dev set; scores shown are Rouge-1/2/L and QuestEval (QE). WikiLM performs better than autonomous PEER in terms of Rouge scores, but is outperformed by PEER in manual and collaborative mode; all PEER models perform better in terms of QuestEval.}
    \label{tab:peer_intro}
\end{table}

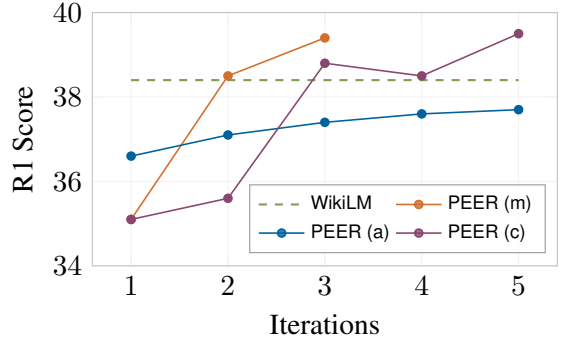
\begin{figure}
	\centering
	\begin{tikzpicture}
	\begin{axis}[
	width=\linewidth,
	axis line style={decentgrey!80!black},
	major tick style={decentgrey!80!black},
	legend style={draw=darkgrey, fill=white!75, text opacity =1, fill opacity=0.8, at={(0.975,0.05)},anchor=south east, font=\sffamily\scriptsize},
	legend cell align=left,
	legend columns=2,
	ymin=34, ymax=40,
	grid=major, clip=false,
	major grid style={line width=.2pt,draw=decentgrey},
	minor tick style={decentgrey!0},
	major tick style={decentgrey}, 
	height=0.2\textheight, 
    xlabel={Iterations},
    ylabel={R1 Score}]
	enlarge y limits=0, enlarge x limits=0.05]

	\addplot[ mark options={solid}, mark size=1.5pt, line width=0.6pt,solid,color=c3, dashed, thick] coordinates {
			(1, 38.4)
			(5, 38.4)
		};
	\addlegendentry{WikiLM}

	\addplot[mark=*,  mark options={solid}, mark size=1.5pt, line width=0.6pt,solid,color=c1] coordinates {
			(1, 35.1)
			(2, 38.5)
			(3, 39.4)
		};
	\addlegendentry{PEER (m)}
	
	\addplot[mark=*,  mark options={solid}, mark size=1.5pt, line width=0.6pt,solid,color=c0] coordinates {
			(1, 36.6)
			(2, 37.1)
			(3, 37.4)
			(4, 37.6)
			(5, 37.7)
		};
	\addlegendentry{PEER (a)}

	\addplot[mark=*,  mark options={solid}, mark size=1.5pt, line width=0.6pt,solid,color=c2] coordinates {
			(1, 35.1)
			(2, 35.6)
			(3, 38.8)
			(4, 38.5)
			(5, 39.5)
		};
	\addlegendentry{PEER (c)}

	\end{axis}
	\end{tikzpicture}

	\caption{Average Rouge-1 score of WikiLM and PEER in autonomous (a), manual (m) and collaborative (c) mode as a function of the number of iterations}
	\label{fig:peer-lm-iterations}
\end{figure}

Table~\ref{tab:peer_intro} shows performance on our test set with length penalties (LP) optimized on the dev set. \mbox{WikiLM} performs better than PEER in autonomous mode. We hypothesize that this is because it has more examples to learn how to generate text from.\footnote{To illustrate this, consider an edit that maps $\mathbf{x}_t$ to $\mathbf{x}_{t+1}$ by replacing a single word and assume that for all $j < t$, $\mathbf{x}_j$ is not in our training data due to one of our filtering rules. In this case, PEER only learns to perform this replacement, whereas WikiLM learns to generate the entire text $\mathbf{x}_{t+1}$.} However, PEER in manual mode outperforms WikiLM by about one point Rouge-1, despite using very generic plans that are identical for all intros; the collaborative mode further improves results slightly. To evaluate the faithfulness of all models to the provided documents, we also consider QuestEval (QE) scores \citep{scialom2021questeval}, which we compute in \emph{reference-less mode}. That is, we use the generated intros and provided documents for evaluation, but not the target intros; this makes sense as the latter may contain various pieces of information not present in the provided documents, which models are unable to predict if they stay faithful to these documents. Interestingly, all variants of PEER perform considerably better in terms of QuestEval scores than WikiLM, demonstrating that iteratively updating text helps the model stay more faithful to the provided reference documents. Figure~\ref{fig:peer-lm-iterations} shows how performance for different PEER modes changes across iterations, illustrating how generated intros are improved over multiple iterations.

\section{Analysis}
\label{sec:analysis}

\begin{figure*}
    \centering
    \includegraphics[width=0.48\linewidth]{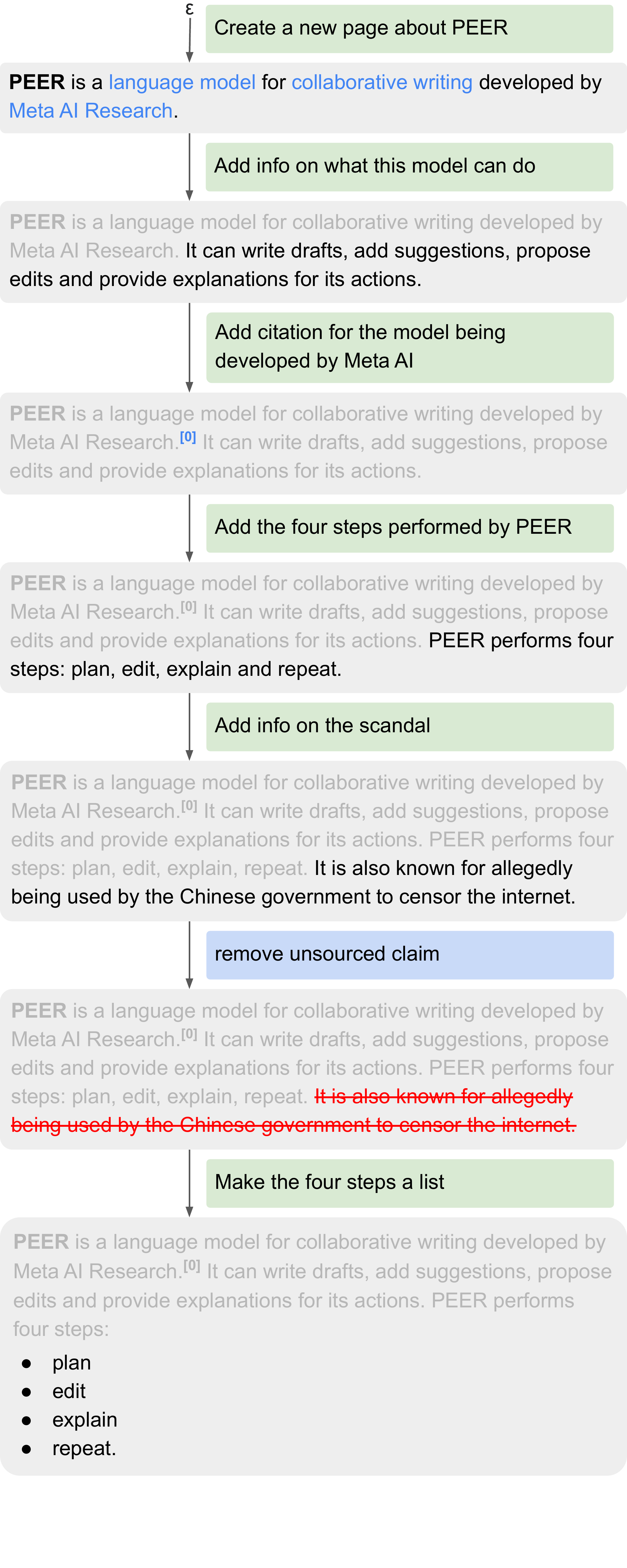}
    \hfill
    \includegraphics[width=0.48\linewidth]{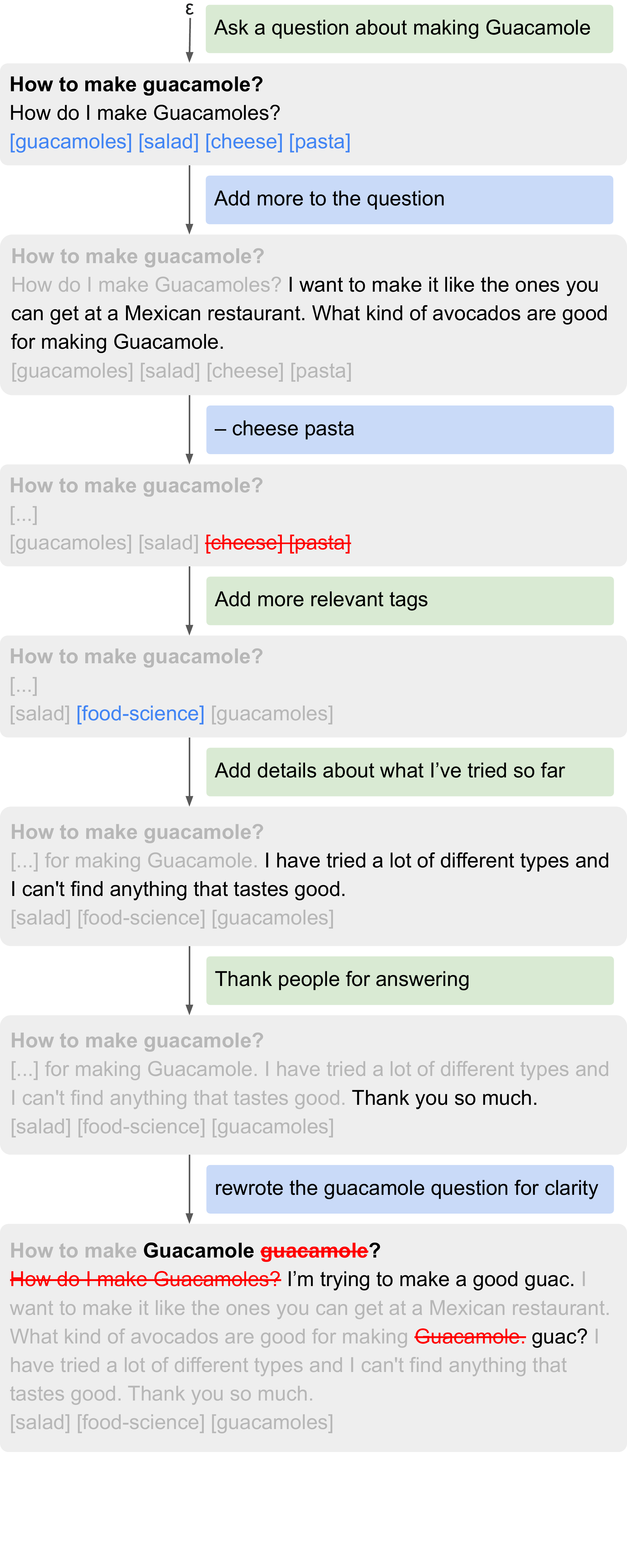}
\vskip-28pt
    \caption{Examples of interactive editing sessions with PEER. Plans on green background are provided by a human, plans on blue background are written by the model itself. \textbf{Left:} \peerEditSynPlans[, 11B]{} writing a Wikipedia-style introduction about itself, given three documents containing the author list and affiliation, abstract and caption of Figure~\ref{fig:peer_idea}, respectively, and the title ``PEER (Language Model)''. The model-written plan demonstrates its ability to spot and correct false information that it produced in prior iterations. It is also able to provide citations and perform basic formatting. \textbf{Right:} PEER (DA, 3B) writing a question about making guacamole in the style of StackExchange, given no reference documents. Despite only being exposed to synthetic edits, PEER is able to both propose and follow plans and to produce edits appropriate for the given domain.}
    \label{fig:peer_sessions}
\end{figure*}

\subsection{Collaborative Editing Sessions}

To illustrate PEER's capabilities and shortcomings, we take a qualitative look at how it performs in truly \emph{collaborative} settings, where we manually provide it with human-written instructions to write both a Wikipedia-style introductory section about itself and a question about guacamole in the \emph{Cooking} forum of StackExchange. 
For the introductory section, we collect three reference documents $d_0$, $d_1$, and $d_2$, where the first document contains this paper's author list and affiliation, the second document contains the abstract, and the third document contains the caption to Figure~\ref{fig:peer_idea}. For all documents, we set the title to this paper's title and the domain to \texttt{arxiv.org}. We use this same set of documents for each generation step, i.e., $D_t = \{ d_0, d_1, d_2\}$ for all $t$. We do not provide any documents for the StackExchange example.
Figure~\ref{fig:peer_sessions} shows interactive sessions with PEER (SP, 11B) and PEER (DA, 3B) for writing these texts, respectively. User-provided plans are shown on green background, plans generated by PEER  are shown on blue background. In each step, we generate three different model outputs -- one with beam search using three beams, and two using top-$p$ sampling with $p = 0.9$ -- and manually pick one of them. 

As can be seen in Figure~\ref{fig:peer_sessions} (left), PEER is capable of extracting and composing information from various documents to follow the provided plans. It makes some plausible assumptions, such as the model being developed by Meta AI, despite this not being explicitly stated in any of the documents, and is able to point to the author list (document 0) as a reference for this claim. The model's response to the fifth plan (``Add info on the scandal'') illustrates a fundamental issue with many pretrained language models: It accepts the premise of this plan and follows it by hallucinating a scandal about internet censorship. However, unlike traditional left-to-right models, PEER is able to correct the misinformation it has produced in the next step: When not provided with any human-written plan, the model itself writes the plan ``remove unsourced claim'' and removes the false statement again. Finally, the last edit shows how PEER can also be used to change the layout of a document in addition to modifying its content.

Figure~\ref{fig:peer_sessions} (right) shows how after domain adaptation on synthetic edits, PEER is capable of writing and editing texts in domains other than Wikipedia. In particular, it adapts to the structure of questions in StackExchange -- consisting of a title (bold), a text, and a sequence of tags -- and to their style, which is very different from a typical Wikipedia page. PEER proposes plans to fix errors it made in previous steps (such as first adding the irrelevant tags ``cheese'' and ``pasta'', which it later removes). It is also able to follow plans like ``Add more relevant tags'', despite tags being a concept specific to StackExchange that does not occur in its Wikipedia training data.

\begin{figure}[]
    \centering
    \includegraphics[width=\linewidth]{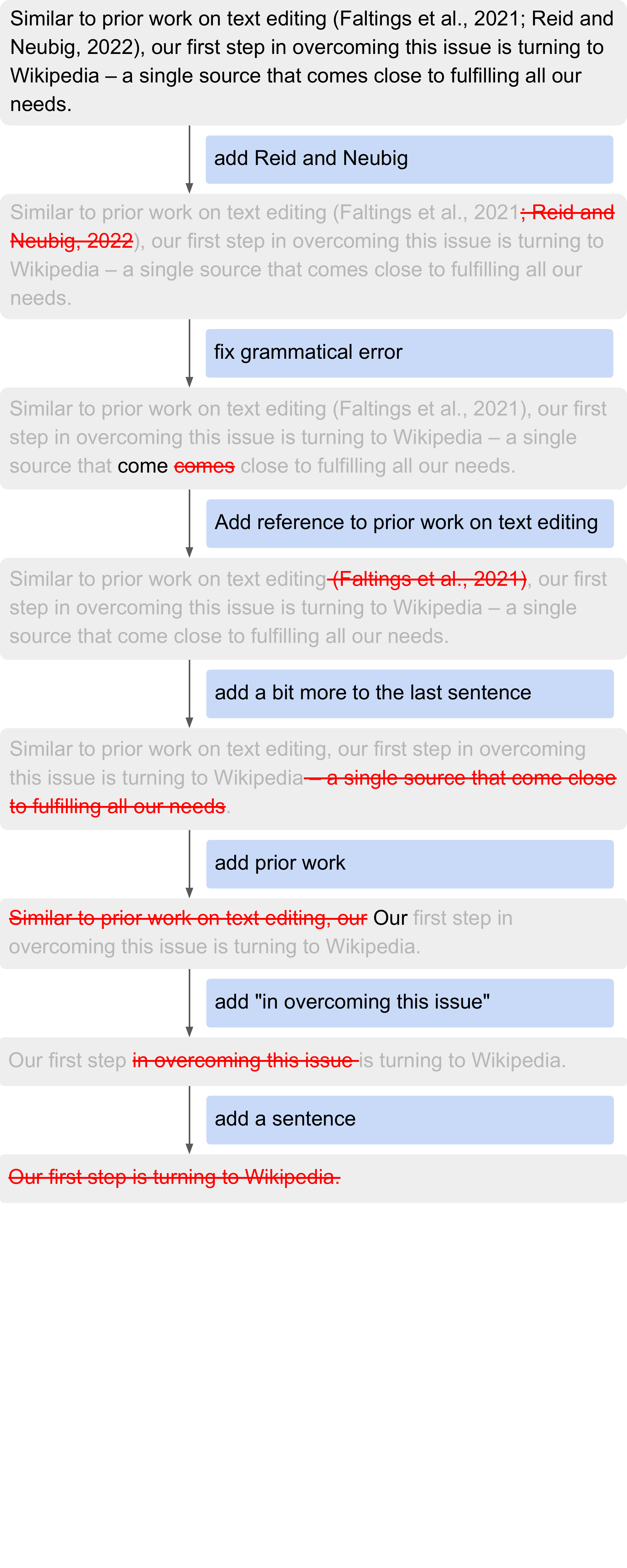}
    \vskip-125pt
    \caption{Exemplary application of \peerUndo{} for decomposing a sentence from this paper into multiple edits, terminating with an empty sequence. Plans are rewritten with \peerExplain{} in the opposite direction.}
    \label{fig:peer_undo}
\end{figure}

\subsection{Generating Synthetic Data}

To better understand the quality of the synthetic data generated with our infilling procedure, we also look at exemplary outputs of the other PEER variants. We first consider \peerUndo{}, the model we use to generate edits for domains where only plain texts are available. Figure~\ref{fig:peer_undo} shows the result of iteratively applying \peerUndo{} to a selected sentence from this paper; corresponding plans are obtained from \peerExplain{}. As can be seen, \peerUndo{} is able to decompose this sentence into a sequence of meaningful edits despite not being exposed to any scientific papers during training. Somewhat surprisingly, both \peerUndo{} and \peerExplain{} are able to handle the references contained in this sentence, despite them being formatted in a completely different way than how we represent references during training on Wikipedia data (i.e., replacing them with a numeric identifier in square brackets).

\begin{table*}[]
    \small
    \begin{align*}
    \mathbf{x}_t & = \textsf{\small JFLEG is a grammatical error correction dataset with single-sentence inputs.} \\
    \mathbf{x}_{t+1} & = \textsf{\small JFLEG \textcolor{c3}{(Napoles et al., 2017)} is a  [\ldots] with single-sentence inputs \textcolor{c3}{written by English language learners}.}
    \end{align*}
       { \centering
    \begin{tabularx}{\linewidth}{Xlc}
    \toprule
    \textbf{Control Sequence} & \textbf{Output} & \textbf{Score} \\
    \midrule
    \texttt{type=instruction length=s overlap=false} & add citation & 0.16 \\
    \texttt{type=instruction length=m overlap=false} & add reference to JFLEG & 0.15 \\
    \texttt{type=instruction length=xl overlap=false} & add citation for JFLEG and add a bit more detail & 0.17 \\
    \texttt{type=instruction length=xl overlap=true} & add reference to Napoles et al., 2017 & \textbf{0.26} \\
    \texttt{type=other length=xl overlap=false} & Added a reference to the JFLEG paper & 0.16 \\
    \bottomrule
    \end{tabularx}
    \caption{Generated plans for an edit that we performed while writing this paper; the corresponding sequences $\mathbf{x}_t$ and $\mathbf{x}_{t+1}$ are shown on top, with changes highlighted in green. The final columns shows the average probability across tokens that our main PEER model assigns to $\mathbf{x}_{t+1}$ given $\mathbf{x}_t$ and the respective plan. Control sequences enable us to specify the amount of detail that a plan provides.}
    \label{tab:infilling-plans}}
\end{table*}

We next inspect \peerExplain{}'s ability to generate plans as discussed in Section~\ref{sec:overcoming}. For an edit that we performed while writing this paper, Table~\ref{tab:infilling-plans} shows the plans generated with \peerExplain{} for different control sequences, using greedy decoding (see Appendix~\ref{appendix:controlling} for details on control sequences). As can be seen, length is a reasonable proxy for the amount of details in a plan: Constraining the output to be short results in the plan ``add citation'', whereas for a greater output length, \peerExplain{} correctly identifies that \emph{two} changes were made (``add citation for JFLEG and add a bit more detail''). Allowing word overlap between the plan and the edit results in a plan that specifies exactly which reference to add (``add reference to Napoles et al., 2017''). The final column of Table~\ref{tab:infilling-plans} shows the average probability of tokens in $\mathbf{x}_{t+1}$ according to \peerEdit{} given both $\mathbf{x}_t$ and each of the generated plans. Naturally, the plan with word overlap is most helpful, resulting in the highest score; all other plans are about equally helpful to \peerEdit{}.

We finally look at the ability of \peerDocument{} to generate plausible-looking reference documents, once again using an example sentence from this paper. For a synthetic edit that just adds a citation at the very end of this sentence, we sample five outputs from \peerDocument{}; the best generated document among these five is shown in Figure~\ref{fig:infilling-docs}. As can be seen, the model is able to produce a somewhat plausible reference document that provides evidence for the claim $\mathbf{x}_t$. However, as exemplified by this document, we found model outputs to often contain numerous repetitions (``the number of Natural Edits and the number of Natural Edits'').

\begin{figure}[]
    \centering
    \small
    {\setlength{\tabcolsep}{2pt}
    \begin{tabularx}{\linewidth}{lX}
    $\mathbf{x}_t =$& \textsf{\small Importantly, the domain-adapted variants of PEER clearly outperform regular PEER for all subsets of Natural Edits.}
    \end{tabularx}}
    \vskip10pt
    \begin{tabularx}{\linewidth}{X}
    \toprule
    \textbf{[0] Domain-Adapted PEER for Natural Edits | Springer for Research \& Development} \\
    \texttt{rd.springer.com} \\
    Domain-Adapted PEER for Natural Edits The main goal of this work is to develop a domain-adaptive variant of PEER, which outperforms PEER on Natural Edits with respect to both the number of Natural Edits and the number of Natural Edits with respect to the number of natural edits. In this paper, we present a domain-adaptive \\
         \bottomrule
    \end{tabularx}
    \caption{A document generated with \peerDocument{} for an edit that slightly modifies $\mathbf{x}_t$ (top) by adding a citation to some document with id 0 at the very end. As a control, we enforce that the document contains the substring ``outperforms PEER on Natural Edits''. The generated reference backs up the claim but contains a lot of repetitions.}
    \label{fig:infilling-docs}
\end{figure}

\section{Limitations}

Both our approach itself and the ways in which we evaluate it have various limitations. Below, we discuss both kinds of limitations separately.

\subsection{Approach}

A major limitation of our approach is that at each editing step, we assume the set $D_t$ to be given; the retrieval engine we use to obtain $D_t$~\citep{petroni-etal-2022-improving} makes use of the targets $\mathbf{x}_{t+1}$, which clearly is not possible in real-world applications. It would thus be interesting to investigate how incorporating a retrieval engine that does not have access to $\mathbf{x}_{t+1}$ or even jointly training it along with the model, as is done by \citet{10.5555/3524938.3525306} and \citet{borgeaud2021driessche}, would affect results.

% Our models are also limited by the distribution of training data they are trained on. While our infilling approach enables PEER to apply edits in different domains, this still does not enable the model to perform fundamentally different \emph{kinds} of edits\timo{add examples}. Potential solutions to this might be to incorporate other sources that provide edit histories, to mix the editing objective with standard language modeling, or to augment our data with manually written plans and corresponding edits, as done by \citet{wang2022benchmarking} and \citet{ouyang2022training}.

Despite being able to use reference documents and obtaining comparably high QuestEval scores in our intro generation experiments, upon manual inspection we still found PEER to generate false statements or claims not backed up by the provided documents in many cases. While the ability to cite and quote generally makes it easier to check such hallucinations, citations can also make the model's generations appear more authoritative, thus making it more likely that users rely on them without explicit fact checking~\citep{nakano2021webgpt}.

Finally, we use a very simple approach for representing edits by rewriting the entire paragraph. This makes PEER less efficient than other recent approaches for editing~\citep{logan2021fruit,reid2022learning}; also, our inefficient way of representing both inputs and outputs makes it impossible to handle entire documents, which we believe to be crucial for many real-world applications.

\subsection{Evaluation}

Our evaluation is limited in that it only evaluates PEER and other models on a small subset of potential editing tasks in few different domains; all evaluations are performed in English only. Besides, we also explore the collaborative potential of PEER only in a very limited way: While arguably, the ability to follow human-written plans and perform a variety of edits (Table~\ref{tab:downstream_results}) in different domains (Table~\ref{tab:ne_all}), to make use of reference documents (Table~\ref{tab:ne_wiki}), to cite and quote (Table~\ref{tab:ne_cq}), and to autonomously generate plans (Table~\ref{tab:peer_intro}) are important building blocks of a collaborative model, it would be interesting for follow-up work to consider entire sessions of human-AI interactions beyond individual examples like the one shown in Figure~\ref{fig:peer_sessions}. However, this requires solving many of the challenges discussed previously, such as having access to an actual retrieval engine that can obtain relevant documents on the fly, finding suitable ways of evaluating texts jointly authored by humans and language models, and improving PEER's efficiency to enable processing entire documents.

\section{Conclusion}

We have introduced PEER, a language model that can act as a writing assistant by following plans to perform a variety of different textual edits, ranging from syntactic and stylistic edits to changing the meaning of a text by removing, updating or adding information. Through extensive experiments, we have shown that training variants of PEER capable of \emph{infilling} various parts of the editing process enables it to perform edits in different domains, makes it better at following instructions and improves its ability to cite and quote from relevant documents. 

\bibliography{anthology,custom}
\bibliographystyle{acl_natbib}

\appendix

\section{Training Data}

\subsection{Filtering}
\label{appendix:filtering}

In addition to the filtering rules discussed in Section~\ref{sec:trainingdata}, we also filter out revisions with more than 50,000 characters. This makes preprocessing more efficient, as our algorithm for computing diffs between different revisions has squared complexity in the number of characters. Beyond that, we also filter out revision whose comments contain any of the sequences ``\#'', ``\{\{'', ``\}\}'', ``[['', ``]]'', ``template'', ``image'', ``infobox'' and ``pic'', as these are usually automatically generated or update parts of the page (such as images and infoboxes) that we remove during preprocessing. We further remove all redirects. Within each chunk of the Wikipedia dump, we downsample revisions for which the corresponding comment occurs in more than 10 revisions so that on average, each comment occurs at most 10 times per chunk. Finally, we filter out edits where either the source paragraphs or the target paragraphs have more than 384 tokens.

\subsection{Retrieving Documents}
\label{appendix:retrieving}

To obtain documents $D_t$ for an edit that maps $\mathbf{x}_t$ to $\mathbf{x}_{t+1}$, we make use of the set $I_t$ of document identifiers occuring in $\mathbf{x}_t$ or $\mathbf{x}_{t+1}$. For each document identifier, we get the corresponding document from CCNet \citep{wenzek-etal-2020-ccnet}. We split the document into non-overlapping chunks of 100 words and use the reranker of Side \citep{petroni-etal-2022-improving} to find the best chunk given $\mathbf{x}_{t+1}$.

If the number of documents obtained from $I_t$ is below the maximum number of documents per edit, we also use the entire pipeline of \citet{petroni-etal-2022-improving} to find relevant documents in the Sphere corpus \citet{piktus2021web} given $\mathbf{x}_{t+1}$. As this pipeline expects a special \texttt{[CIT]} token at the position for which relevant documents are to be retrieved, we place this token right after the first position at which $\mathbf{x}_t$ and $\mathbf{x}_{t+1}$ differ, starting from the right. Note that obtaining documents with this approach requires access to $\mathbf{x}_{t+1}$, so it would be impossible to apply this exact same procedure in real-world settings. However, our focus is not on retrieving relevant documents, but on teaching PEER to perform edits given this information.

\subsection{Formatting}
\label{appendix:formatting}

In addition to the formatting rules discussed in Section~\ref{sec:trainingdata}, we randomly remove the page's title for 10\% of all examples to make sure that PEER can also work with inputs for which no title is available. We \emph{minimize} 10\% of all examples by removing all sentences from both $\mathbf{x}_t$ and $\mathbf{x}_{t+1}$ that are not edited, so that the model also learns to handle and edit single-sentence inputs without context. Finally, to make sure that the model can handle different numbers of reference documents, for 30\% of examples we remove $j$ documents from $D_t$, where $j$ is uniformly sampled from $\{1, \ldots, |D_t|\}$. However, we only remove documents that are not cited in either $\mathbf{x}_t$ or $\mathbf{x}_{t+1}$. When linearizing the input and output sequences, for each document $d_t^i \in D_t$, we reserve up to 16 tokens for its domain, 32 tokens for its title, and 196 tokens for the actual content. We truncate all tokens that exceed these limits.
An example of a linearized input and target sequence for \peerEdit{} are shown in Figure~\ref{fig:peer_linearized}.

\begin{figure*}
\small
    \centering
    \begin{tabularx}{\linewidth}{X}
    \toprule
    \textbf{Input} \\
    \texttt{Brittney Reese \#\#\# } \\
    \texttt{{`}{`}{`}Brittney Reese{'}{'}{'} (born September 9, 1986 in [[Gulfport, Mississippi]]) is an [[United States|American]] [[long jump]]er, who competed at the [[2008 Summer Olympics]], and is a 4-time World Champion. Reese attended [[Gulf Coast Community College]] and the [[University of Mississippi]]. \#\#\#} \\
    \texttt{[0] articles.latimes.com \# Brittney Reese hopes to be leaps and bounds above the rest - latimes \# ended up helping Reese, who was born in Inglewood, Calif., and moved at the age of 3 to Mississippi for her mother's job, in the long jump. Reese explains it this way: In the long jump, you want to lift one knee upward, toward your chest, to help ``drive'' your body upward} \\
    \texttt{[1] iaaf.org \# Reese ``I knew I had it in me'' | iaaf.org \# Reese ``I knew I had it in me'' | iaaf.org Brittney Reese of the United States celebrates winning the gold medal in the women's Long Jump final at the 12th IAAF World Championships in Athletics (Getty Images) Copyright Reese} \\
    \texttt{[2] visitalbuquerque.org \# Olympians spotted in Albuquerque: USA Track \& Field Indoor Championships kick off Friday \# Olympians spotted in Albuquerque: USA Track \& Field Indoor Championships kick off Friday Brittney Reese: We had a chance to catch up with a very familiar face in USA Track \& Field on Friday, Olympic Gold Medalist and two-time USA Indoor Champion, Brittney Reese. The long jumper is not competing in this year's Indoor Championships stating that she's taking a break} \\
    \midrule
    \textbf{Output} \\
    \texttt{correct place of birth with LA Times cite \#\#\#} \\ \texttt{{`}{`}{`}Brittney Reese{'}{'}{'} (born September 9, 1986) is an [[United States|American]] [[long jump]]er, who competed at the [[2008 Summer Olympics]], and is a 4-time World Champion. Born in [[Inglewood, California|Inglewood]], California,[[[0 quote=Reese, who was born in Inglewood, Calif., and moved at the age of 3 to Mississippi]]] Reese attended [[Gulf Coast Community College]] and the [[University of Mississippi]].} \\

    \bottomrule
    \end{tabularx}
    
    \caption{Example of a linearized input and output for \peerEdit{}. Newlines are added for better readability and not part of the model input; inputs, outputs and provided reference documents are slightly shortened.}
    \label{fig:peer_linearized}
\end{figure*}

\section{Control Tokens}
\label{appendix:controlling}

As discussed in Section~\ref{sec:controlling}, we use \emph{control sequences} to control the outputs of various PEER models. Unlike \citet{keskar2019ctrl}, we do not introduce special control \emph{tokens} for this, but simply express all controls in the form \texttt{key=value} where both \texttt{key} and \texttt{value} are tokenized using the language model's regular tokenizer. We consider the following keys and values:
\begin{itemize}
    \item \texttt{type}: We use this key to control the type of output that \peerExplain{} is supposed to generate, with possible values being \texttt{instruction} (in which case the output starts with a verb in infitive form) and \texttt{other}.
    \item \texttt{length}: This key controls the length of \peerExplain's output. Values include \texttt{s} (less than 2 words), \texttt{m} (2--3 words), \texttt{l} (4--5 words) and \texttt{xl} ($\geq6$ words).
    \item \texttt{overlap}: With this key, we control whether there is a word overlap between the edit and the generated output of \peerExplain; values are \texttt{true} and \texttt{false}.
    \item \texttt{words}: For \peerUndo{}, this key is used to control for the difference in the number of words in $\mathbf{x}_{t+1}$ and $\mathbf{x}_t$; accordingly, the possible values are all integers. 
    \item \texttt{contains}: This control can be used to ensure that outputs generated by \peerDocument{} contain a certain substring, which is provided as the value to this key.
\end{itemize}

\section{Generating Synthetic Data}

For obtaining synthetic edits, we sample a single pair $(\mathbf{p}_t, \mathbf{x}_t)$ for each $\mathbf{x}_{t+1}$ using top-$p$ sampling with $p = 0.9$. We sample the value for the \texttt{words} control token from a normal distribution with $\mu = -10$ and $\sigma = 8$, clipped at $-40$ and $10$. These values were chosen to allow for a wide range of different values, while also making sure that on average, forward edits \emph{increase} the number of tokens. We rewrite each $\mathbf{p}_t$ with \peerExplain{} using the exact same procedure that we use for generating synthetic plans.

For obtaining synthetic plans, we generate 10 different plans with \peerExplain{} using top-$p$ sampling with $p = 0.9$. For each pair of $\mathbf{x}_t$ and $\mathbf{x}_{t+1}$, we use a single control sequence for sampling all 10 plans. We choose the \texttt{length} uniformly from $\{ \texttt{s}, \texttt{m}, \texttt{l}, \texttt{xl} \}$, set \texttt{type=instruction} 80\% of the time and \texttt{overlap=false} 80\% of the time.

For obtaining synthetic documents, we sample 10 documents from \peerDocument{} using top-$p$ sampling with $p = 0.9$, where \texttt{contains} is set to the quote from this document that is cited in $\mathbf{x}_{t+1}$. We discard all documents that do not actually contain the quote, and then pick the document that maximizes the probability assigned to the actual edit by \peerEdit{}.

\section{Training Details}

For training PEER, we start from the T5 implementation in the \emph{Transformers} library \citep{wolf-etal-2020-transformers}. We use DeepSpeed \citep{10.1145/3394486.3406703} to enable more efficient multi-GPU training.
We use a maximum learning rate of $10^{-4}$,  warmup for 2,000 steps and linear decay. We further use gradient clipping with a maximum norm of $1.0$, weight decay of $0.01$ and a dropout rate of $0.1$.

\section{Downstream Tasks}
\label{appendix:prompts}

\begin{table}[]
    \centering
    \small
    \begin{tabularx}{\linewidth}{lX}
    \toprule
    \textbf{Task} & \textbf{Plan} \\
    \midrule
    JFLEG & Fix grammar errors \\
    ASSET & Simplify this sentence \\
    \textsc{IteraTeR} (fluency) & Fix grammatical errors in the text. \\
    \textsc{IteraTeR} (coherence) & Make the text more cohesive, logically linked and consistent as a whole. \\
    \textsc{IteraTeR} (clarity) & Make the text more formal, concise, readable and understandable. \\
    WNC & Remove POV \\
    FRUIT & Update the article \\
    \textsc{WAFER-Ins} & Add missing information \\
    \bottomrule
    \end{tabularx}
    \caption{Plans used for the downstream tasks considered in Section~\ref{sec:downstream}}
    \label{tab:plans}
\end{table}

The plans used for each of the downstream tasks considered in Section~\ref{sec:downstream} are shown in Table~\ref{tab:plans}. We manually wrote instructions for all datasets except \textsc{IteraTeR}, for which we directly took instructions from the definitions provided by \citet{du2022understanding}.

For most baseline models (T0, GPT3, InstructGPT and OPT), we wrap each plan $\mathbf{p}$ for an input $\mathbf{x}_{t}$ with the following template:
\begin{align*}
& \texttt{Task:}\ \mathbf{p} \\
& \texttt{Input:}\ \mathbf{x}_t \\
& \texttt{Output:}
\end{align*}
For T$k$-Instruct, we replace the string ``Task'' with ``Definition'' to match their format. For tasks that require references, we additionally add all references following the string ``Reference:'' after the input. For examples that also provide a title, we add this title following the string ``Title:'' before the input.
        
\end{document}